\documentclass[10pt,twocolumn,letterpaper]{article}

\usepackage{iccv}
\usepackage{times}
\usepackage{epsfig}
\usepackage{graphicx}
\usepackage{float}
\usepackage{caption}
\usepackage{amsmath}
\usepackage{amssymb}
\usepackage{pifont}
\usepackage{booktabs}
\usepackage{array}
\usepackage{arydshln}
\usepackage{multirow}
\usepackage{siunitx}
\usepackage{lipsum}
\usepackage{rotating}
\usepackage{csquotes}
\usepackage{placeins}

\usepackage{subcaption}
\usepackage[pagebackref=true,breaklinks=true,letterpaper=true,colorlinks,bookmarks=false]{hyperref}

\iccvfinalcopy 

\newcommand{\parheader}{\smallskip\noindent\textbf}

\newcommand{\xmark}{\ding{55}}%

\def\oursA{\textit{In-Distribution}}
\def\oursB{\textit{Out-of-Distribution}}

\begin{document}

\title{Reassessing the Limitations of CNN Methods for Camera Pose Regression}

\author{Tony Ng$^{1}\thanks{Equal contribution. Order decided by a coin-flip.}$ \hspace{6pt}Adrian Lopez-Rodriguez$^{1}\footnotemark[1]$ \hspace{6pt}Vassileios Balntas$^{2}$   \hspace{6pt}
Krystian Mikolajczyk$^{1}$\\
$^1$Imperial College London \\
$^2$Facebook Reality Labs \\
{\tt\small \{tony.ng14, adrian.lopez-rodriguez15, k.mikolajczyk\}@imperial.ac.uk}\\
{\tt\small vassileios@fb.com}}

\maketitle

\begin{abstract}
   In this paper, we address the problem of camera pose estimation in outdoor and indoor scenarios. In comparison to the currently top-performing methods that rely on 2D to 3D matching, we propose a model that can directly regress the camera pose from images with significantly higher accuracy than existing methods of the same class. We first analyse why regression methods are still behind the state-of-the-art, and we bridge the performance gap with our new approach. Specifically, we propose a way to overcome the biased training data by a novel training technique, which generates poses guided by a probability distribution from the training set for synthesising new training views.
   Lastly, we evaluate our approach on two widely used benchmarks and show that it achieves significantly improved performance compared to prior regression-based methods, retrieval techniques as well as 3D pipelines with local feature matching. 
\end{abstract}

\vspace{-5pt}
\section{Introduction}
Visual localisation is the problem of localising an image \wrt to a known and pre-mapped environment.
This could be in the form of coarse place recognition~\cite{arandjelovic2016netvlad,torii201524-7} or predicting the accurate 6DoF camera pose~\cite{kendall2015posenet,sattler2018benchmarking} that includes both rotational and positional estimations. 
The latter is especially important for enabling mass deployment of technologies in robotics, autonomous driving and spatial AR.

Traditionally, localisation methods relied on local features~\cite{dusmanu2019d2net,lowe2004sift,revaud2019r2d2,tian2020hynet}, 3D point clouds computed using Structure from Motion (SfM)~\cite{sarlin2018hfnet,sattler2017active,schonberger2016colmap}, and solving for the pose with minimal solvers~\cite{gao2003p3p} in a RANSAC~\cite{fischler1981ransac}.
These methods have up to now dominated the relevant benchmarks, and are able to predict highly accurate 6DoF pose~\cite{sattler2018benchmarking,sattler2017are-large-scale-3D} but are costly in both time and memory.
With the advent of CNNs, the authors of PoseNet~\cite{kendall2015posenet} demonstrated that it is possible to directly predict 6DoF camera position in a known scene by end-to-end learning with a convolutional network. 
Many works on pose regression soon followed suit~\cite{balntas2018relocnet,ding2019camnet,kendall2017posenet-geometric,saha2018anchornet,walch2017posenet-lstm} due to the appeal of fast inference, no need for 3D models during inference, and increased potential of deployment in  different devices compared to traditional 3D methods. 

\begin{figure}[!t]
    \vspace{0pt}
    \begin{center}
        \includegraphics[width=\linewidth, trim={
        0pt, 5pt, 0pt, 5pt}, clip]{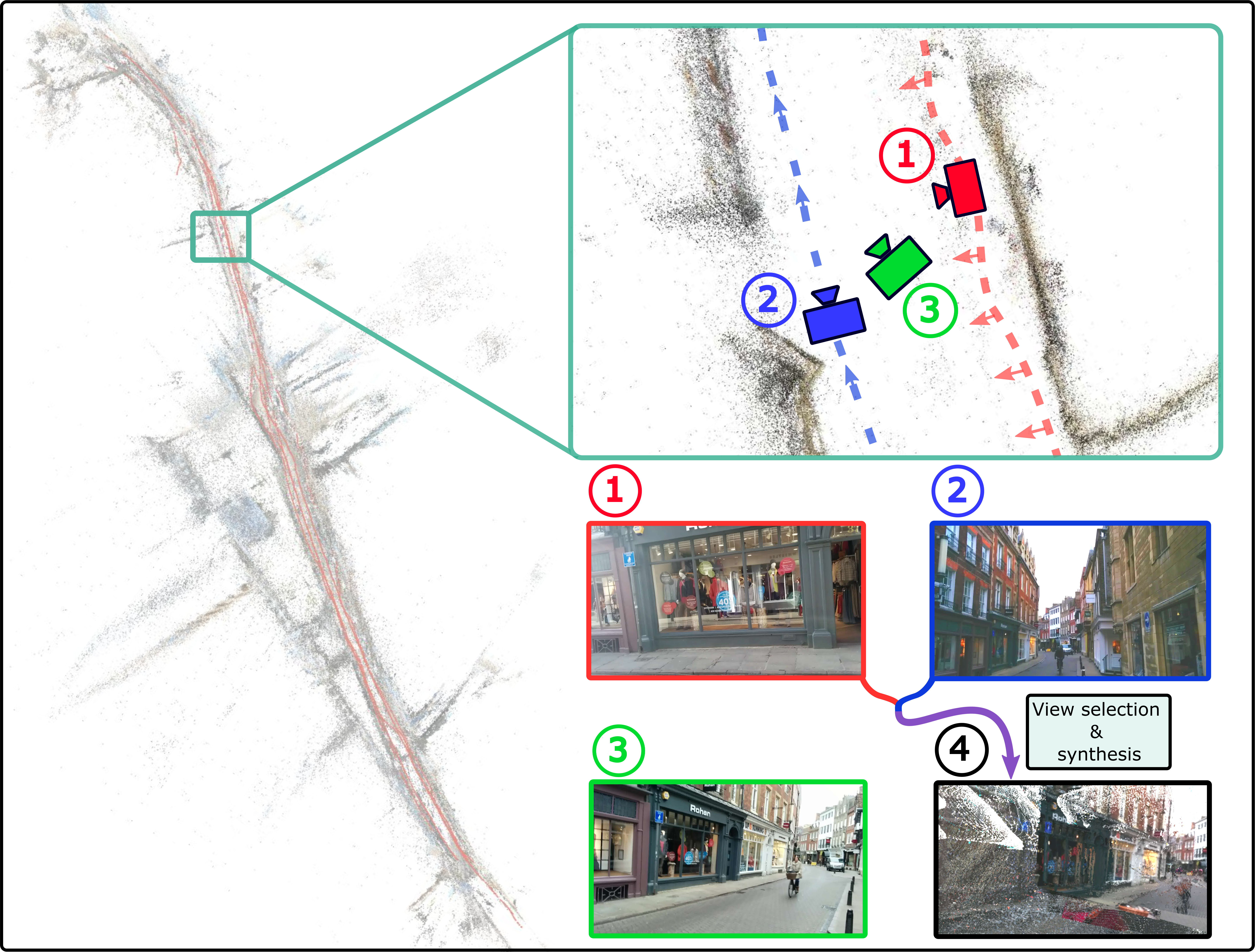}
    \end{center}
    \vspace{-15pt}
    \caption{Challenge facing CNN methods for camera pose regression. We show a bird's-eye view of the 3D model of a street. The \textcolor{red}{\textbf{(1)}} and \textcolor{blue}{\textbf{(2)}} dotted lines and arrows are the trajectories and orientations of two separate training sequences. As the training data is heavily biased towards specific views, CNN methods do not extrapolate well to the test pose in \textcolor{green}{\textbf{(3)}}. Our pose regression method alleviates this by learning from \textbf{(4)}~synthesised views sampled from an unbiased pose distribution.}
    \label{fig:teaser}
    \vspace{-12pt}
\end{figure}

%
Despite the initial success, pose regression methods have been on the back foot since the study by Sattler \etal~\cite{sattler2019understanding} showed that the performance of pose regression methods is closer to less accurate image retrieval~\cite{pion2020benchmarking} than to 3D structure-based methods~\cite{sattler2017active}. 
This is due to the fact that learning-based methods do not extrapolate well beyond the poses they encounter in training. This problem is especially important in the context of visual localisation~\cite{glocker20137scenes,kendall2015posenet,sattler2018benchmarking}, since the test data is often sampled during a rather structured acquisition process, e.g. a pedestrian walking down a road in a straight continuous trajectory~\cite{kendall2015posenet} or a driving car~\cite{maddern2017robotcar} showing little variations in camera orientation.
This particular bias is illustrated in Figure~\ref{fig:teaser} where we observe that in a scene where existing learning-based methods fail to learn or localise (\textit{Street}~\cite{kendall2015posenet}), the camera poses are biased towards particular viewpoints and orientations.
Although prior works attribute this to the CNN's incapacity to learn large and complex scene structures~\cite{brachmann2018dsac++,sattler2019understanding}, we argue that the network is capable of achieving high accuracy if unbiased data is available.
One way to address this issue would be to increase the size of the training data with random views. However, this is costly and does not guarantee to remove the bias. Instead we focus on generating synthetic images with a probability-guided process with the aim to reduce the gap between training and test distributions.
%



Our main contributions can be elaborated in the context of the observations in Sattler \etal, 2019~\cite{sattler2019understanding} that can be summarised by the following quote:
\vspace{-5pt}
\begin{displayquote}
\textit{``Pose regression techniques
are often closer in performance to image retrieval than to
structure-based methods.... Even with
large amounts of training data, pose regression does not
reach the same performance as structure-based methods.''}\\
\end{displayquote}
\vspace{-15pt}
Specifically: 
\noindent{\bf a)} We show that regression methods can be much closer or even better than structured-based methods in accuracy. In addition, we are the first to propose a regression method that outperforms Active Search~\cite{sattler2017active} in one of the scenes of the Cambridge Landmarks~\cite{kendall2015posenet} benchmark. \\ 
\noindent{\bf b)} We propose a novel pipeline for training camera pose regression methods based on a probability-driven selection and generation of synthetic views. We also propose a novel way of densely sampling depths to improve the quality of synthetic images.\\
\noindent{\bf c)} We show that a transformer~\cite{vaswani2017transformer} based architecture is more suitable for inferring relative pose regression, outperforming the standard MLP approaches adopted by all previous methods. \\ 
\noindent{\bf d)} We show how to make use of view synthesis to improve localisation performance, which has not been demonstrated simultaneously for both indoor and outdoor benchmarks. \\
\noindent{\bf e)} Our approach is the first regression method to improve upon image retrieval on the largest and most complex \textit{Street} scene from~\cite{kendall2015posenet}.\\
\vspace{-15pt}
\section{Related work}
\label{sec:related-work}
In this section we briefly discuss the related techniques in the area of visual localisation and view synthesis. 
\subsection{Visual localisation}
\label{subsec:related-localisation}

\parheader{Image retrieval}
approximates the camera pose estimation problem by matching an unseen query to images with known poses. 
This however only gives a coarse estimate and is often used for general place recognition~\cite{arandjelovic2016netvlad,torii201524-7}. 
Nevertheless, retrieval methods are scalable~\cite{arandjelovic2014dislocation,sattler2016large-scale-burst} and are robust across varying conditions~\cite{arandjelovic2016netvlad,sattler2017are-large-scale-3D}.
Therefore, they are often used as the first step to hierarchical localisation~\cite{humenberger2020kapture,sarlin2018hfnet} and relative pose regression~\cite{ding2019camnet,saha2018anchornet}. A recent review of retrieval based localisation can be found in~\cite{pion2020benchmarking}. 

\parheader{Structure-based} 
methods rely on 3D models, e.g. by SLAM~\cite{mur-artal2016orb-slam,newcombe2011dtam} or SfM~\cite{schonberger2016colmap,sweeney2015theia}.
Local features~\cite{dusmanu2019d2net,lowe2004sift,noh2017delf,revaud2019r2d2} are extracted and matched across images to establish 2D-2D correspondences, 
which are then lifted to 2D-3D~\cite{sattler2017active} point-cloud. 
Minimal solvers such as PnP~\cite{gao2003p3p} are then employed within a RANSAC-loop~\cite{fischler1981ransac} to predict the camera pose. 
Although structure-based methods are the most accurate~\cite{jin2020image,sattler2018benchmarking}, they have limited scalability due to the complexity of feature matching and 3D reconstructions.

\parheader{Absolute pose regression} was  proposed with 
PoseNet~\cite{kendall2015posenet} by directly regressing camera poses from CNN features~\cite{simonyan2015VGG,he2016ResNet}.
These methods do not require an explicit 3D model, 
as they learn the 3D scene implicitly through data.
Further improvements have been made by exploring geometric constraints~\cite{brahmbhatt2018mapnet,kendall2017posenet-geometric}, uncertainty~\cite{kendall2016uncertainty} and sequences~\cite{walch2017posenet-lstm,xue2020multi-view}. 
It has been shown in~\cite{sattler2019understanding} that absolute pose regression models are closer to image retrieval than to structure-based models in terms of performance, as they overfit to the training data.

\parheader{Relative pose regression} 
~\cite{balntas2018relocnet,ding2019camnet,saha2018anchornet} also works without 3D models, but relies on CNN features to predict the relative poses between images. At test time, image retrieval is first used to find the coarse pose prediction and combined with the relative one to find the absolute pose.
Methods for relative pose regression usually outperform the absolute ones~\cite{sattler2019understanding}, as the distribution between training and testing sets are closer for relative poses~\cite{chidlovskii2020adversarial}.

\parheader{Scene coordinates regression} are hybrids of structure-based and regression methods. 
The static~\cite{brachmann2018dsac,brachmann2018dsac++,shotton2013scene} or temporal~\cite{zhou2020kfnet} 3D \textit{scene coordinates} are learnt to establish dense 2D-3D correspondences, which are then used to predict the pose with minimal solvers~\cite{fischler1981ransac,gao2003p3p}.
Scene coordinate regression methods are very accurate at small scales, but fail to converge on larger scenes. In addition, they require a significant amount of time for both the training and the inference stages.

\subsection{View synthesis}
\label{subsec:view-synthesis}
There are two main categories of methods for novel view synthesis.
The first is based on surface modelling, which explicitly exploits the geometry \eg with depth maps or surface meshes~\cite{extremeview,niklaus20193dkensburn}, to generate an image by projecting the 3D structure onto the novel view.
They generalise to a variety of views, and while normally suffer from low levels of photo-realism, blending techniques can greatly improve the quality of images \cite{riegler2020free}.
The second category is volumetric and rendering based, 
which densely represent the scene's appearance implicitly with techniques such as multi-plane images~\cite{srinivasan2019pushing} or Neural Radiance Fields (NeRF)~\cite{mildenhall2020nerf}.
They produce highly photo-realistic images, but do not generalise to large viewpoint changes.
Recent improvements on modelling uncertainties \cite{martinbrualla2021nerf-in-the-wild} made it possible to generalise to challenging scenes. 
Nevertheless, volumetric methods require significant amount of training, are more complex and much slower than surface modelling methods despite recent efforts to improve the speed of convergence~\cite{liu2020neural}. 

For visual localisation, the use of synthetic views has been explored in place recognition~\cite{torii201524-7}, pose verification~\cite{taira2018inloc, zhang2020aachenv_1_1} and pose estimation~\cite{naseer2017deep, sattler2019understanding, wan2020boosting}.
Unlike the naive data augmentations in \cite{brachmann2018dsac, li2020hierarchical, li2018full} which include only in-plane transformations,
\cite{naseer2017deep, wan2020boosting} leverage 3D using a depth prediction network. However, this limits the range of transformations possible around existing poses for absolute pose regression, making it closer to the data augmentation approach. 
As part of the study, Sattler \etal~\cite{sattler2019understanding} investigated grid-based pose densification using multi-view stereo (MVS) in a small-scale experiment on a single scene, showing the potential of view synthesis in improving both image retrieval and absolute pose regression.
\subsection{Transformers in visual localisation}
\label{subsec:transfomer}
The Transformer~\cite{vaswani2017transformer}  has been broadly adopted by the vision community in various tasks, but its performance still needs to be further investigated. The self-attention mechanism has been recently shown to be beneficial for image retrieval~\cite{ng2020solar} and absolute pose regression when applied to feature maps~\cite{wang2020atloc}.
SuperGlue~\cite{sarlin20superglue} applies self and cross-attentions for better feature matching in a structure-based approach, achieving state-of-the-art in several localisation benchmarks~\cite{jin2020image,sattler2018benchmarking}. 

With attention being the central component of the Transformer, these results are promising for the potential of Transformers in relative pose regression. This motivates the examination of its architecture in the context of this work.
%
\section{Method}
\label{sec:method}
In this section, we present the main components of our novel localisation approach for a pinhole camera model. 
First, we discuss how we select and generate synthetic views. We then show how to make use of these synthetic images for training a Transformer-based relative pose regression model.

\subsection{Preliminaries}\label{subsec:synth-views}
\label{subsec:preliminaries}

\parheader{Projecting to a novel view with a depth map.}
Given that an RGB image $\mathcal{I} \in \mathbb{R}^{h,w,3}$ and its depth map $\mathcal{D} \in \mathbb{R}^{h,w}$ exist in a known 3D environment, \ie we have both the camera intrinsic $\mathbf{K}$ and extrinsic $\mathbf{C}$ matrices from 3D reconstruction, we are able to obtain the synthetic image from a novel view $\mathcal{\hat{I}}$ with three steps. 
\textbf{1.)} For a pixel $\mathbf{p} = (u,\,v)^\top$ in $\mathcal{I}$, its 3D point in world coordinate $\mathbf{x} = (x, \, y, \, z)^\top$ is obtained by $\mathbf{x}(\mathbf{p}) = \mathbf{C} \mathcal{D}(\mathbf{p}) \mathbf{K}^{-1}\mathbf{\tilde{p}}$, where $\mathbf{\tilde{p}} = (u,\,v,\, 1)^\top$ denotes the homogeneous vector. 
\textbf{2.)} We denote the extrinsic matrix of the novel view $\mathbf{\hat{C}} = T(\mathbf{C})$, where $T \in \mathbb{S}\mathbb{E}(3)$ is the transformation from the source view $\mathbf{C}$ to our target novel view $\mathbf{\hat{C}}$. 
\textbf{3.)} Lastly, we re-project the 3D point $\mathbf{x}$ to image plane of $\mathcal{\hat{I}}$ by $\mathbf{\hat{p}} = \pi (\mathbf{\hat{x}})$, where $\mathbf{\hat{x}} = (\hat{x},\,\hat{y},\,\hat{z})^\top = \mathbf{K} \mathbf{\hat{C}}^{-1} \mathbf{x}$, and $\pi$ denotes the dehomogenisation $\pi(\mathbf{\hat{x}}) = \left(\lfloor \hat{x}/\hat{z} \rfloor , \, \lfloor \hat{y}/\hat{z} \rfloor \right)^\top$.
We compact the function that returns the new depth value(s)  
%
${\hat{z}}$ at this point as  $\boldsymbol{\xi}_{\textit{proj}}(\mathbf{p})|_{\mathcal{D}, T} := \{{\hat{z}} \in \mathbf{\hat{X}}_z\,|\,\pi(\mathbf{\hat{X}}) = \mathbf{\hat{p}} \}$, the target image $\mathcal{\hat{I}}$ is hence formed by retrieving all corresponding RGB values from $\mathcal{I}$,
\begin{equation}
\label{eq:project_to_image_plane}
    \mathcal{\hat{I}} (\mathbf{\hat{p}}) = 
        \begin{cases}
        \mathcal{I} \left( \underset{\mathbf{p}}{\arg\min}~ \boldsymbol{\xi}_\textit{proj}\left(\mathbf{p}\right)|_{\mathcal{D}, T} \right), & | \hat{z} | > 0 \\[10pt]
        (1, 1, 1)^\top, & \hat{z}  = \emptyset.
        \end{cases}
\end{equation}
The first condition in Equation~\ref{eq:project_to_image_plane} ensures that occlusions are retained, 
and the second condition is that holes from warping the depth maps are filled with a white background.
\begin{figure}[!t]
    \vspace{-10pt}
    \begin{center}
        \includegraphics[width=1.\linewidth, trim={
        5pt, 5pt, 0pt, 0pt}, clip]{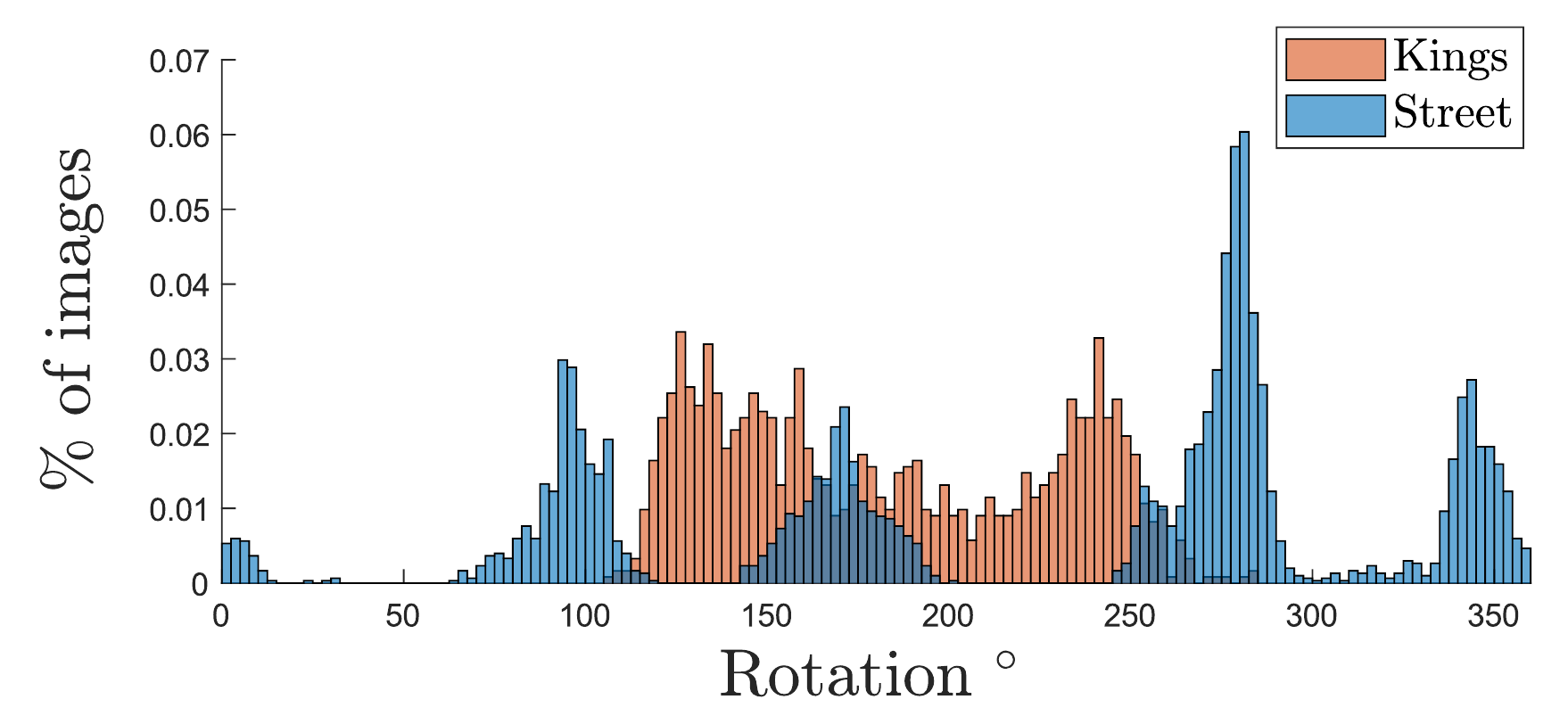}
    \end{center}
    \vspace{-15pt}
    \caption{Distribution of absolute \textit{yaw} rotations of \textit{Kings} \vs \textit{Street} in the Cambridge Landmarks~\cite{kendall2015posenet} dataset. The four peaks in \textit{Street} show there may exist large bias in the data collection process compared to the more uniform \textit{Kings} distribution.}
    \label{fig:histogram}
    \vspace{-5pt}
\end{figure}
\parheader{Motivation - the infamous \textit{Street} scene.}
Recently, Sattler \etal~\cite{sattler2019understanding} demonstrated that absolute pose regression methods perform closer to image retrieval than 3D-based methods.
It is also noted that no regression model so far has been able to produce any meaningful result on the largest scene in Cambridge Landmarks~\cite{kendall2015posenet} - the \textit{Street} scene. 
In fact, the network used in DSAC++~\cite{brachmann2018dsac++} for scene coordinates regression, which achieves state-of-the-art across all other scenes, does not even converge in \textit{Street}.
Also, no current regression method, either absolute or relative, consistently improves upon image retrieval on this scene.
Although some previous works have speculated the scene's size and complexity being the main reason behind this~\cite{brachmann2018dsac++,sattler2019understanding}, these claims are not backed up experimentally.
In the following, we show that this view is not correct, and the culprit is in fact a biased distribution of training poses instead.

\begin{figure*}[!ht]
    \vspace{0pt}
    \begin{center}
        \includegraphics[width=1.\linewidth, trim={
        0pt, 10pt, 0pt, 10pt}, clip]{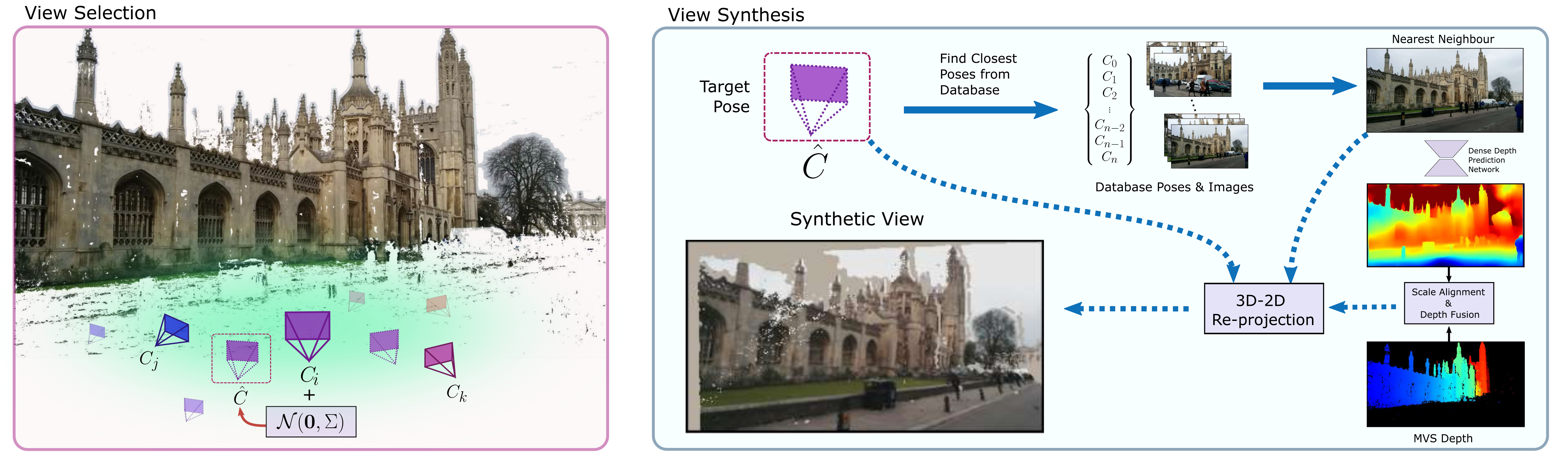}
    \end{center}
    \vspace{-15pt}
    \caption{Illustration of view selection and synthesis for training. \textbf{Left.} We first sample a target pose $\hat{C}$ by a Gaussian (opaque green cloud) based on an existing pose $C_{i}$. \textbf{Right.} We produce synthetic images from $\hat{C}$ by combining densely-predicted but scale-agnostic depths with the sparse metric MVS depths of the nearest images, and re-projecting to the target pose.}
    \label{fig:synthesis}
    \vspace{-8pt}
\end{figure*}

This is illustrated in Figure~\ref{fig:histogram}, where we compare the distributions of camera angles about the \textit{yaw} axis (rotating `left-to-right' from a standing person's perspective) between the training sets of the \textit{Kings} and the \textit{Street} scenes in Cambridge Landmarks~\cite{kendall2015posenet} dataset.
We observe that while the rotations in \textit{Kings} form a flat distribution, there may be a bias within \textit{Street} training distribution showing distinct multiple modes with no angles in between. 
There are four peaks at $90^\circ$ intervals, which suggests that the training images come from four sequences, each facing one predominant direction. 
Intuitively, for a camera pose, such data may violate the inductive bias that allows regression methods to work in other scenes, which is that the training distribution should approximate the test distribution.
Without access to the test data, our best assumption is that the test poses should be uniformly distributed. 
Hence, the bias problem becomes apparent, since in both absolute and relative pose training for \textit{Street}, the network only learns from poses in 90$^\circ$ intervals.
Therefore, we cannot expect the network to be able to extrapolate far beyond such a narrow range of training poses.
To further experimentally address the question 
\textit{`Is Street too large and too complex for a network to learn, or is it learnable if we fix the bias in the data?'}, 
we conduct a simple sanity check.
We first use image retrieval to find the nearest neighbour in the training set for each test query. 
We then overfit the relative poses of these test image pairs to our \textit{Pose Regression Module} from Section~\ref{subsec:localisation_pipeline} to serve as an indicator of the upper-bound accuracy.
With the \textit{Views Selection} and \textit{Synthesis} modules from Section~\ref{subsec:view_selection_and_synthesis}, we produce synthetic views from the training images projected onto the test poses 
%
%
%
and train the \textit{Pose Regression Module} with these synthetic views and their nearest neighbour. Finally, we predict the poses for the test images and their nearest neighbour in training data.
The results are presented in Table~\ref{tab:street_sanity_check}. %
Given that the upper-bound results reduce the median errors from retrieval nearly ten-fold, is it evident that the scale/complexity of the scene is not the main issue. 
Training with known synthetic poses also reduces the error by half compared to the coarse prediction by retrieval.
This suggests that the network is able to learn on synthetic views and improve its generalisation from training data, even in large-scale scenes like \textit{Street} which covers more than \textit{$\SI{50000}{m^2}$}~\cite{kendall2015posenet}. 
%
%
\begin{table}[t!]
\renewcommand{\arraystretch}{1.3}
\addtolength{\tabcolsep}{0.2em}
\begin{center}
        \resizebox{.475\textwidth}{!}{%
        \begin{tabular}{r|l}
            Experiment & Median Error        \\
            \hline\hline
            Real views from test poses (Upper-bound) &  $\SI{0.89}{m}$, $\SI{1.50}{^\circ}$          \\
            Synthetic views projected onto test poses & $\SI{3.54}{m}, \SI{13.9}{^\circ}$ \\
            \hline
            Image retrieval with GeM~\cite{radenovic2018gem} & \SI{6.42}{m}, 23.6$^\circ$
        \end{tabular}
        }
    \vspace{-10pt}
    \caption{
        Sanity check on \textit{Street}. For the first and second experiments we train on test views with real images and synthetic views generated with training images respectively. The last row is the coarse prediction with image retrieval.
        }
    \vspace{-20pt}
\label{tab:street_sanity_check}
\end{center}
\end{table}
\subsection{View selection and synthesis}
\label{subsec:view_selection_and_synthesis}
Given our hypothesis that the distribution of training poses is a potential factor for improving pose regression, we introduce in this section a novel way of acquiring synthetic views for the densification of the training distribution/alignment with test distribution. 
In contrast to \cite{naseer2017deep, wan2020boosting, sattler2019understanding}, we propose a probabilistic sampling method for relative pose regression, which works for multiple scenes, both indoor and outdoor, and is capable of generating pairs of views covering a larger set of poses due to our view-synthesis strategy using MVS depths with multi-image accumulation.
There are two components - \textit{View Selection} and \textit{View Synthesis}, which are illustrated in Figure~\ref{fig:synthesis}.
%
\begin{figure*}[!ht]
    \vspace{-10pt}
    \begin{center}
        \includegraphics[width=1.\linewidth, trim={
        0pt, 5pt, 0pt, 10pt}, clip]{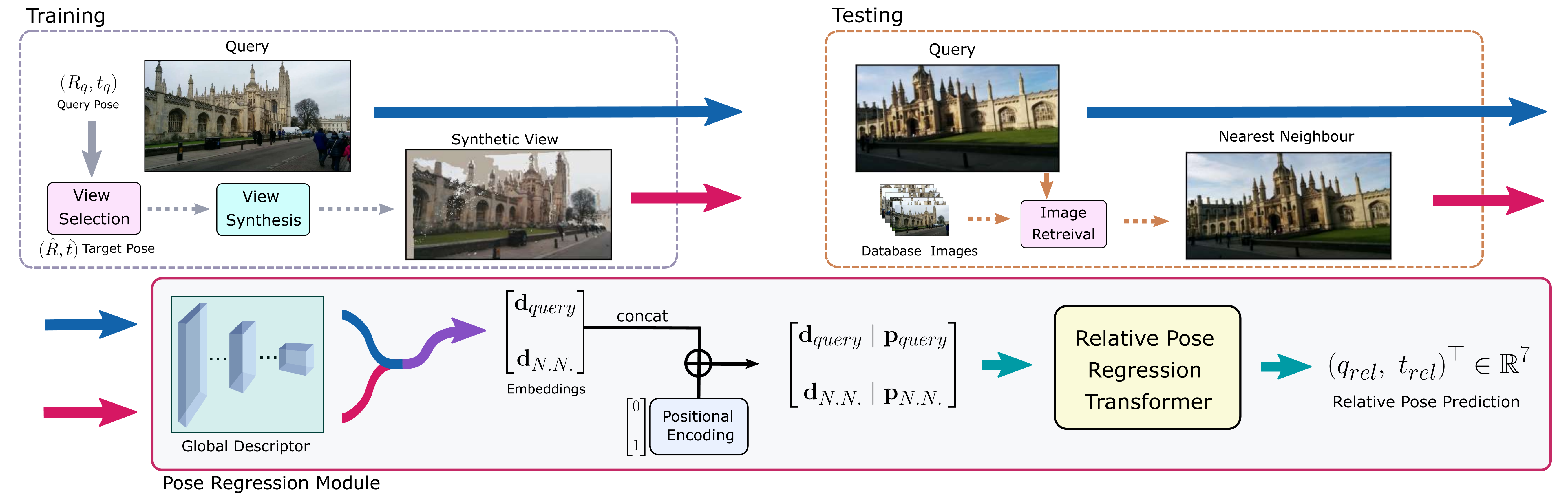}
    \end{center}
    \vspace{-15pt}
    \caption{Our localisation pipeline. The image pair - query and nearest neighbour (N.N.) - are represented by \textbf{\textcolor{blue}{blue}} and \textbf{\textcolor{red}{red}} arrows. During training, the N.N. is synthesised from the process described in Section~\ref{subsec:view_selection_and_synthesis}. In inference, N.N. is determined by an image retrieval method. The image pair is fed into our \textit{Pose Regression Module}, which contains a Transformer that predicts the relative pose between the two.}
    \label{fig:pipeline}
    \vspace{-10pt}
\end{figure*}
%

\noindent\textbf{View Selection.}
To generate our synthetic views, we select camera poses based on a multi-variate Gaussian distribution $\mathcal{N} (\mathbf{0}, \Sigma)$ around the existing poses $\{C_i, C_j, C_k, ...\}$. Using Gaussian increases the probability that the new pose will be close to the existing one thus have sufficient overlap with real images needed for synthesis.
The Gaussian density is visualised by the opaque green cloud on the left of Figure~\ref{fig:synthesis}.
The goal of this component is to select a target pose $\hat{C}$ that alleviates the bias in the training data demonstrated earlier.
This is illustrated in Figure~\ref{fig:synthesis} left again, where $\hat{C}$ is ideally beyond the interpolation $C_i$ and $C_j$ to achieve the filling of missing poses in the overall distribution. 
We propose two variants of this selection process based on our assumptions on the data - \oursA~ and \oursB.
\oursA:~We assume that the inductive bias holds, \ie the training distribution approximates the test distribution. For each training image, we obtain a set of nearest neighbours (N.N.s) using a retrieval method, in our case GeM~\cite{radenovic2016gem}, which we use to form our training query-N.N. pairs.
The target pose $\hat{C} = \left(\hat{q}, \hat{t} \right)^\top \in \mathbb{R}^7$ is obtained using the nearest neighbour's pose $C_{N.N.}$ as a basis
\begin{equation}
    \hat{C} = C_{N.N.} + \mathcal{N}\left(\textbf{0}, \ \Sigma_{in}\right)  
\end{equation}
where $\Sigma_{in} = \begin{bmatrix}
\alpha_q I_{4} & \mathbf{0}\\ 
\mathbf{0} & \alpha_t I_{3}
\end{bmatrix}$ and $\alpha_q, \alpha_t$ are scalar multipliers for the quaternion and translation vectors that aim to add small scale perturbations around $C_{N.N.}$. After the perturbation, the rotation quaternion is normalised to 1.

\oursB:~We assume that a large bias might be present such as in the \textit{Street} example from Figure~\ref{fig:histogram}. 
In this case, we cannot base our synthesised views only on the existing training poses as the network cannot extrapolate out of those ranges. 
Instead, we generate new poses by taking each training image and applying to its pose a large perturbation compared to \oursA. This perturbation is parametrised by the Gaussian $\mathcal{N}(\textbf{0}, \ \Sigma_{out})$ around the training query poses, where $\Sigma_{out}$ is determined from the training data. 
For example, for outdoor scenes such as Cambridge~\cite{kendall2015posenet}, we observe that most images are viewed from roughly the same height and the camera elevation does not vary as much compared to movements on the ground plane. 
Therefore, the scale of perturbations along the two ground plane axes would be larger than the vertical.
For camera translations, this might result in poses where no 3D-points are visible, \eg the camera centre might end up inside a wall. 
Thus, we check within a spatial neighbourhood if there are existing camera poses and use that with an additional small perturbation, as the target translation $\hat{t}$. 
For the target rotation $\hat{q}$, we keep what is selected from adding $\mathcal{N}(\textbf{0}, \ \Sigma_{out})$ to the query rotation.

\parheader{View Synthesis.} 
This is illustrated on the right of Figure~\ref{fig:synthesis}. For each of the training images, we obtain a depth map from a Multi-View Stereo (MVS) process. The resulting MVS depth can be sparse, which affects the quality of our synthetic images. We therefore fuse the MVS depth map with a densely predicted depth using MiDaS~\cite{ranftl2020midas}. Since the MiDaS predicted depth maps do not have a consistent scale, we align its scale with ground-truth sparse MVS depths by following a similar scale and shift alignment method used in the MiDaS paper~\cite{ranftl2020midas}.
We then fill the missing depths (black regions in the `MVS Depth' photo in Figure~\ref{fig:synthesis}) with the now scale-aligned dense depth prediction.
For the \oursA~approach, we expect our target poses to be close to the poses in the training pairs. Hence, we directly project the N.N. to the new pose using the fused depth with the process described in Section~\ref{subsec:preliminaries}.
However, in \oursB, the new synthetic pose can be distant to the training pose it is generated from. We therefore iteratively re-project 3D points according to the pose distance of the source image to the target pose, \ie the vector $(1,1,1)^\top$ in the second condition of Equation~\ref{eq:project_to_image_plane} is replaced by the corresponding RGB values from the projection process of the next closest image.
This is repeated until the image is filled to a certain threshold or a maximum number of iterations.

\subsection{Localisation pipeline} 
\label{subsec:localisation_pipeline}
Our full visual localisation pipeline is illustrated in Figure~\ref{fig:pipeline}. 
The \textit{Pose Regression Module} takes two images, including the query (\textcolor{blue}{blue} arrows) and N.N. (\textcolor{red}{red} arrows), as input in both training and testing. 

In training, we sample synthetic views using the process described in Section~\ref{subsec:view_selection_and_synthesis} and output the synthesised view as the nearest neighbour (N.N.) image. 
At test time, the top match from image retrieval is used as the N.N. image.

Within the \textit{Pose Regression Module}, we first process the images by a global descriptor network, \ie the feature maps from a CNN such as ResNet~\cite{he2016ResNet} plus global average pooling to obtain embeddings $(\mathbf{d}_{query},\ \mathbf{d}_{N.N.})^\top \in \mathbb{R}^{2,c}$, where $c$ is the feature size. Inspired by recent success~\cite{dosovitskiy2021an}, we use a Transformer encoder~\cite{vaswani2017transformer} as our relative pose regression module. We validate experimentally the superiority of the Transformer against a MLP architecture in Section~\ref{sec:ablation_study}.
A learnt positional encoding with the input $(0, 1)^\top \rightarrow (\textbf{p}_{query}, \textbf{p}_{N.N.})^\top \in \mathbb{R}^{2,d}$ is concatenated with the embeddings to form the input vectors $\in \mathbb{R}^{2, c+d}$ to the \textit{Relative Pose Regression Transformer}.
The outputs of the Transformer are the relative rotation in unit quaternion $q_{rel} \in \mathbb{R}^4$ and relative translation $t_{rel} \in \mathbb{R}^3$. 

To localise the query, we first convert quaternion $q_{rel}$ to rotation matrix $R_{rel} \in \mathbb{SO}(3)$. 
The absolute pose prediction is given by $R = R_{N.N.}R_{rel}^\top$ and $t = t_{N.N.} + t_{rel}$, where $(R_{N.N.}, \ t_{N.N.})$ is the pose of the nearest neighbour.

\subsection{Loss function}
\label{subsec:loss_function}
We use a weighted $l_1$-loss to supervise the relative pose predictions
\begin{equation}
    \mathcal{L} = \left \| q_{rel} - q_{rel}^* \right \|_1 e^{\beta} +  \left \| t_{rel} - t_{rel}^* \right \|_1 e^{\gamma}, 
    \label{eq:loss}
\end{equation}
where $q_{rel}^* $ is the relative unit quaternion between the query image and the nearest neighbour converted from the rotation matrix $R^*_{rel} = {R^{*}}^{\top} R_{N.N.}$ and the relative translation $t^*_{rel} = t^* - t_{N.N.}$. 
($R^*$,\ $t^*$) denotes the ground-truth pose of the query.
$\beta, \gamma$ are constants for weighting between the rotation and translation terms.

\begin{table*}[t!]
\scriptsize{
\setlength{\tabcolsep}{3pt}
\renewcommand{\arraystretch}{1.1}
\begin{center}
\vspace{-10pt}
\begin{tabular}{cl|c|c|c|c|c||c|c|c|c|c|c|c}
& & \multicolumn{5}{c||}{\footnotesize{Cambridge Landmarks}} & \multicolumn{7}{c}{\footnotesize{7 Scenes}} \\
 & & Kings & Old & Shop & St. Mary's & Street & Chess & Fire & Heads & Office & Pumpkin & Kitchen & Stairs  \\ \hline

\hline

\rule{0pt}{.9\normalbaselineskip}
\multirow{3}{*}{\begin{sideways} \textcolor{red}{IR} \end{sideways}} & DenseVLAD~\cite{torii201524-7} & 2.80/5.72  & 4.01/7.13 & 1.11/7.61 & 2.31/8.00 & \textcolor{red}{5.16}/\textcolor{red}{23.5} & 0.21/12.5 & \textcolor{red}{0.33}/13.8 & 0.15/14.9 & 0.28/11.2 & 0.31/11.3 & 0.30/12.3 & 0.25/15.8 \\

& DenseVLAD + Inter.~\cite{sattler2019understanding} & \textcolor{red}{1.48}/\textcolor{red}{4.45} & \textcolor{red}{2.68}/\textcolor{red}{4.63} & \textcolor{red}{0.90}/\textcolor{red}{4.32} & \textcolor{red}{1.62}/\textcolor{red}{6.06} & 15.4/25.7 & \textcolor{red}{0.18}/\textcolor{red}{10.0} & \textcolor{red}{0.33}/\textcolor{red}{12.4} & \textcolor{red}{0.14}/\textcolor{red}{14.3} & \textcolor{red}{0.25}/\textcolor{red}{10.1} & \textcolor{red}{0.26}/\textcolor{red}{9.42} & \textcolor{red}{0.27}/\textcolor{red}{11.1} & \textcolor{red}{0.24}/14.7\\ %
& GeM~\cite{radenovic2018gem} & 3.22/5.71 & 4.46/7.11 & 1.57/8.25 & 3.01/9.37 & 6.39/23.6 & 0.27/13.1 & 0.39/15.8 & 0.21/15.7 & 0.37/11.7 & 0.43/11.6 & 0.42/13.4 & 0.34/\textcolor{red}{14.2} \\ \hline

\multirow{2}{*}{\begin{sideways} \textcolor{blue}{3D} \end{sideways}} & Active Search~\cite{sattler2017active} & 0.42/0.55 & 0.44/1.01 & 0.12/0.40 & 0.19/0.54 & \textcolor{blue}{0.85}/\textcolor{blue}{0.8} & 0.04/1.96 & 0.03/1.53 & 0.02/1.45 & 0.09/3.61 & 0.08/3.10 & 0.07/3.37 & \textcolor{blue}{0.03}/\textcolor{blue}{2.22}\\

& DSAC++~\cite{brachmann2018dsac++} & \textcolor{blue}{0.18}/\textcolor{blue}{0.3} & \textcolor{blue}{0.20}/\textcolor{blue}{0.3} & \textcolor{blue}{0.06}/\textcolor{blue}{0.3} & \textcolor{blue}{0.13}/\textcolor{blue}{0.4} & \xmark & \textcolor{blue}{0.02}/\textcolor{blue}{0.5} & \textcolor{blue}{0.02}/\textcolor{blue}{0.9} & \textcolor{blue}{0.01}/\textcolor{blue}{0.8} & \textcolor{blue}{0.03}/\textcolor{blue}{0.7} & \textcolor{blue}{0.04}/\textcolor{blue}{1.1} & \textcolor{blue}{0.04}/\textcolor{blue}{1.1} & 0.09/2.6\\

\hline\hline
\multirow{5}{*}{\begin{sideways} APR \end{sideways}}
& PoseNet~\cite{kendall2015posenet} & {1.92}/{5.40} & 2.31/{5.38} & {1.46}/{8.08} & {2.65}/{8.48} & - & {0.32}/8.12 & {0.47}/{14.4} & {0.29}/12.0 & {0.48}/7.68 & {0.47}/8.42 & {0.59}/8.64 & {0.47}/13.8 \\

& Geometric PoseNet~\cite{kendall2017posenet-geometric} & 0.88/1.04 & {3.20}/3.29 & 0.88/3.78 & 1.57/3.32 & {20.3}/{25.5} & 0.13/4.48 & 0.27/11.3 & {0.17}/13.0 & 0.19/5.55 & {0.26}/4.75 & 0.23/5.35 & {0.35}/12.4\\

& PoseNet LSTM~\cite{walch2017posenet-lstm} & 0.99/3.65 & 1.51/4.29 & {1.18}/{7.44} & 1.52/{6.68} & \xmark & {0.24}/5.77 & {0.34}/11.9 & {0.21}/13.7 & {0.30}/8.08 & {0.33}/7.00 & {0.37}/8.83 & {0.40}/13.7\\

 
& MapNet~\cite{brahmbhatt2018mapnet} & 1.07/1.89 & 1.94/3.91 & {1.49}/4.22 & {2.00}/4.53 & - & {0.08}/3.25 & 0.27/11.7 & {0.18}/13.3 & {0.17}/5.15 & {0.22}/{4.02} & {0.23}/4.93 & {0.30}/12.1  \\

& GRNet~\cite{xue2020multi-view} & {0.59}/{0.65} & 1.88/{2.78} & {0.50}/{2.87} & 1.90/{3.29} &  {14.7}/{22.4} & {0.08}/{2.82} & {0.26}/{8.94} & {0.17}/{11.4} & 0.18/{5.08} & {0.15}/{2.77} & 0.25/{4.48} & {0.23}/{8.87} \\
\hline
\multirow{5}{*}{\begin{sideways} RPR \end{sideways}} 
  
\rule{0pt}{.9\normalbaselineskip}
& RelocNet~\cite{balntas2018relocnet} (SN) & - & - & - & - & - & {0.21}/10.9  &  \textcolor{black}{0.32}/11.8  &  {0.15}/13.4  &  {0.31}/10.3  & 0.40/{10.9}  &  {0.33}/10.3 &  {0.33}/11.4\\

& RelocNet~\cite{balntas2018relocnet} (7S) & - & - & - & - & - & 0.12/4.14 & 0.26/10.4 & {0.14}/10.5 & 0.18/5.32 & {0.26}/4.17 & 0.23/5.08 & {0.28}/7.53\\ 

& AnchorNet~\cite{saha2018anchornet} & 0.57/0.88 & 1.21/2.55 & 0.52/2.27 & 1.04/2.69 & {7.9}/24.2 & 0.06/3.89 & 0.15/10.3 & 0.08/10.9 & 0.09/5.15 & 0.10/2.97 & 0.08/4.68 & 0.10/9.26\\ 

& CamNet~\cite{ding2019camnet} & - & - & -  & - & - & 0.04/1.73 & \textbf{0.03}/1.74 & 0.05/\textbf{1.98} & \textbf{0.04}/1.62 & \textbf{0.04}/1.64 & \textbf{0.04}/1.63 & \textbf{0.04}/1.51 \\
\cdashline{2-14}[4pt/1.61803398875pt]
\rule{0pt}{.9\normalbaselineskip}
& Ours - \oursA & \textbf{0.24}/\textbf{0.38} & 0.83/1.05 & \textbf{0.26}/\textbf{1.06} & 0.58/1.56 & 5.81/16.2 & 0.04/1.69 & 0.13/5.22 & 0.08/5.95 & 0.09/2.70 & 0.12/2.48 & 0.11/2.93 & 0.26/5.13 \\
& Ours - \oursB & 0.35/0.51 & \textbf{0.65}/\textbf{0.93} & 0.32/0.94 & \textbf{0.43}/\textbf{1.18}& \textbf{3.85}/\textbf{7.18} & \textbf{0.03}/\textbf{0.92} & 0.04/\textbf{1.32} & \textbf{0.04}/2.78 & \textbf{0.04}/\textbf{1.35} & \textbf{0.04}/\textbf{0.98} & \textbf{0.04}/\textbf{1.45} & 0.07/\textbf{1.44} 
 
\end{tabular}
\end{center}
}%
\vspace{-15pt}
\caption{Comparison of visual localisation methods on the \textit{Cambridge Landmarks}~\cite{kendall2015posenet} and \textit{7 Scenes}~\cite{glocker20137scenes} datasets.  We report the median errors in translation (m) / rotation ($^\circ$). The four types of methods are image retrieval (IR), structure-based (3D), absolute pose regression (APR) and relative pose regression (RPR), which our method belongs to. We highlight the best results for IR, 3D and all regression methods with \textcolor{red}{red}, \textcolor{blue}{blue} and \textbf{bold}. [(-) no results reported  (\xmark) method fails to train.]}%
\label{tab:results_main}
\vspace{-10pt}
\end{table*}
\subsection{Implementation details}
\label{sec:implementation_details}

\parheader{View selection and synthesis.} 
For our \oursA~view selection in outdoor datasets, we set $\alpha_q=0.02$, $\alpha_t=\SI{1}{m}$, whereas we set  $\alpha_q=0.02$, $\alpha_t=\SI{0.1}{m}$ for our indoor datasets.
For the \oursB~approach,  we give a detailed explanation of the scales used for the view selection in the supplementary material.

\parheader{Relative regression module.} 
The global descriptor backbone is ResNet34~\cite{he2016ResNet}, and we input the descriptor in our \textit{Relative Pose Regression Transformer}, which is 6 layers deep, each layer containing 8 heads. We concatenate a positional encoding term to both the query and nearest neighbour global descriptors before inputting them to the transformer.
The detailed architecture is presented in the supplementary material.

\parheader{Training.}
For each scene, we train the \textit{Pose Regression Module} with the loss in Equation~\ref{eq:loss} for 1200 epochs with batch size of 16. We optimise using Adam~\cite{kingma2015Adam} at an initial learning rate of $10^{-4}$, with a decay rate of $0.1$ every 500 epochs. Our code is implemented using PyTorch and we use Kornia for our view-synthesis pipeline~\cite{riba2020kornia}. We also mix real pairs during training by sampling for each query from the top GeM~\cite{radenovic2018gem} N.N.s (see supplementary material for more details about real pairs sampling).
Images are resized to 180 pixels on the shorter side and random colour jitter is applied. 
For the loss in Equation~\ref{eq:loss}, $(\beta, \gamma)$ are set to $(3,0)$ for outdoor and $(0,0)$ for indoor scenes.

\parheader{Inference.}
For each test query, we retrieve the nearest neighbour from the training set with GeM. Images are resized to 180 pixels on the shorter side but without colour jitter. The inference speed is over 300 images/s on a Titan RTX with a batch size of 16, which is high, mainly due to the small size of the images we use. Detailed timing comparisons with other methods are shown in the supplementary material.
\section{Results}
\label{sec:results}

In this section, we present and discuss the visual localisation results on the two main benchmarks for pose regression: \textit{Cambridge Landmarks}~\cite{kendall2015posenet} and \textit{7 Scenes}~\cite{glocker20137scenes}. 
%
\subsection{Datasets} 
\parheader{Cambridge Landmarks}~\cite{kendall2015posenet} is a collection of five outdoor scenes taken by a mobile phone. Each scene depicts a prominent landmark, with a wide variety of viewpoint and illumination changes, as well as dynamic objects.
The exception is the largest \textit{Street} scene, which contains sequences along a $\SI{500}{m}$-long street and the training images are taken around four specific angles as shown in Section~\ref{subsec:preliminaries}. For this dataset, we generate MVS depth maps in metric scale using COLMAP and the Kapture~\cite{kapture2020} pipeline with AP-GeM~\cite{revaud2019aploss} and R2D2~\cite{revaud2019r2d2}.

\parheader{7 Scenes}~\cite{glocker20137scenes} is the common benchmark for indoor pose regression methods.
It contains seven indoor scenes taken by a Kinect camera. Unlike \textit{Cambridge Landmarks}, camera trajectories in \textit{7 Scenes} are less structured and include more elevation changes. 
Although there are no dynamic objects, a lot of objects in \textit{7 Scenes} are poorly textured. The given Kinect depth is not aligned with the RGB information, so we use the depth information generated from a 3D model provided by past works \cite{brachmann2020dsacstar}.

\subsection{Comparison with state-of-the-art} %
We compare the performance of our visual localisation pipeline with existing methods in Table~\ref{tab:results_main}.
As Sattler \etal~\cite{sattler2019understanding} pointed out, both datasets are challenging for pose regression methods as the training and test poses are sampled from separate sequences and the trajectories differ significantly. 
Following \cite{sattler2019understanding}, we include all four major types of localisation methods.
These include image retrieval (IR), structure-based (3D), absolute pose regression (APR) and relative pose regression (RPR).

From Table~\ref{tab:results_main}, our method gives the lowest median errors, by a large margin, when compared to other existing pose regression methods, including APR and RPR. 
On the \textit{Kings} scene in \textit{Cambridge Landmarks}~\cite{kendall2015posenet}, our \oursA~variant  gives ($\SI{0.24}{m}/\SI{0.38}{^\circ}$) which outperforms even Active Search~\cite{sattler2017active} ($\SI{0.42}{m}/\SI{0.55}{^\circ}$), which was a long-standing point of reference when comparing APR and RPR methods' validity against 3D and IR. It also  comes very close to DSAC++~\cite{brachmann2018dsac++} ($\SI{0.18}{m}, \SI{0.3}{^\circ}$).
To our best knowledge, this the first pose regression method that outperforms \textit{Active Search} on a \textit{Cambridge Landmarks} scene.

On \textit{Street}, a scene where no previous regression methods have improved upon image retrieval, our  \oursA~and \oursB~  improve upon GeM~\cite{radenovic2018gem} retrieval, which we use in our pipeline, and as upon the DenseVLAD retrieval \cite{sattler2019understanding,torii201524-7}.

The results for \oursB~are significantly better than for all other regression methods, reducing the median errors in translation and rotation by 45\% and 60\%, respectively when compared to the state-of-the-art in RPR AnchorNet~\cite{saha2018anchornet}.
The median errors of \oursB~on \textit{Old} and \textit{St. Mary's} are even lower than that of \oursA. 
These two scenes are the more challenging ones amongst the smaller scenes in \textit{Cambridge Landmarks}~\cite{kendall2015posenet}, as suggested by their relatively higher median errors in all existing regression methods when compared to \textit{Kings} and \textit{Shop}.
%
\begin{figure*}[!t]
    \vspace{-10pt}
    \begin{center}
        \includegraphics[width=1.\linewidth, clip]{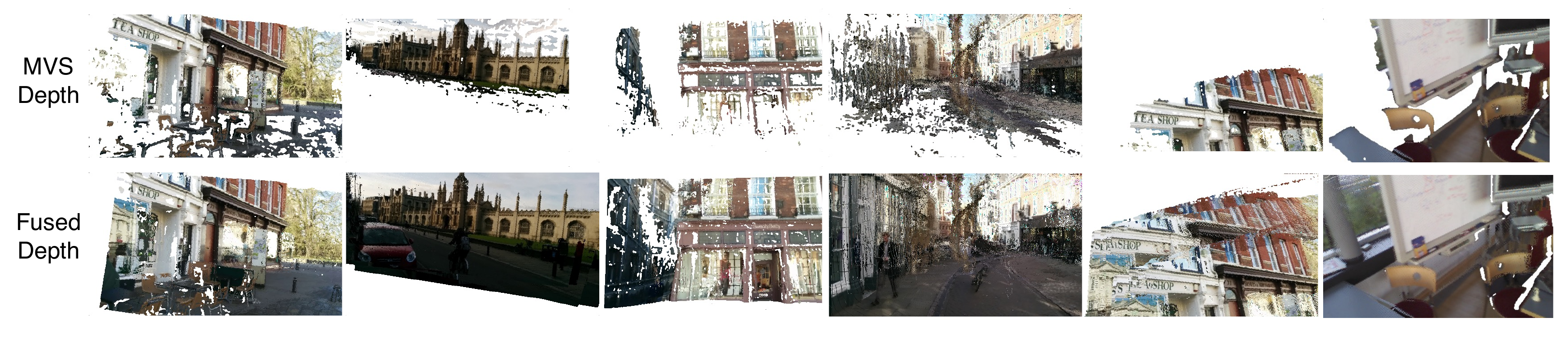}
    \end{center}
    \vspace{-23pt}
    \caption{Examples of novel views synthesised with our approach. Using only the \textit{MVS Depth} limits the quality of the synthesised image due to the depth sparsity. Our \textit{Fused Depth}, presented in Section~\ref{subsec:view_selection_and_synthesis}, allows for denser synthetic views.}
    \label{fig:synth_images}
    \vspace{-5pt}
\end{figure*}

These results show that our \oursB~approach for view selection and synthesis generalises well to all scales and compositions of scenes. More importantly, it is robust to biases in the data, such as in \textit{Street}.
On the other hand, the \oursA~approach is suitable when the training distribution is more aligned with the test distribution, such as in \textit{Kings} and \textit{Shop}.

On \textit{7 Scenes}, both our \oursA~and especially \oursB~method again outperform most regression methods significantly.
Comparing to the best indoor RPR method CamNet~\cite{ding2019camnet}, our \oursB~results match in translation errors and mostly outperform in rotations.
However, it is to be noted that CamNet, additionally from using scene-specific retrieval, uses multiple branches for coarse-to-fine pose predictions and predictions from multiple nearest neighbours, while our methods only require a single query-N.N. prediction at inference.

We compare the average median errors on \textit{Cambridge Landmarks} scenes (with the exception of \textit{Street}) of \oursA~($\SI{0.48}{m}/\SI{1.01}{^\circ}$) and \oursB~($\SI{0.44}{m}/\SI{0.89}{^\circ}$) to Active Search~\cite{sattler2017active} ($\SI{0.29}{m}/\SI{0.63}{^\circ}$) and DenseVLAD + Interpolation~\cite{sattler2019understanding} ($\SI{1.67}{m}/\SI{4.87}{^\circ}$).
The gap between the median errors of our methods and that of Active Search ($\approx\SI{0.2}{m}/\SI{0.4}{^\circ}$) is 6-10 times less than that to DenseVLAD + Interpolation ($\approx\SI{1.2}{m}/\SI{4}{^\circ}$). 
Similar observations are made from the results for \textit{7 Scenes}.
This validates our approach and  supports our claim that {our regression method is the first to  achieve a performance much closer to structure-based methods than to image retrieval.}

\section{Ablation studies}

\label{sec:ablation_study}
In this section, we analyse the respective impact of the three main components of our approach. 

\parheader{Training with real vs. synthetic views.} 
\label{subsec:training_with_real_vs_synthetic_views}
From Table~\ref{tab:ablation_main}, we observe that the median errors in \textit{Cambridge Landmarks} are significantly reduced by about 50\% when training with synthetic views by \oursA~sampling compared to using only real images. 
In order to verify whether the gain in performance is solely due to the increased number of training samples, or if the use of synthesised views truly aligns the training and test distributions, we conduct an experiment which is presented in Table~\ref{tab:ablation_proportion}.
We choose two scenes from \textit{7 Scenes} with a different number of training images. For each scene, we sample a ratio of training images, which we use to generate a single synthetic view each, using our \oursB~method to form our training pairs. 
\begin{table}[t!]
\begin{center}
\setlength{\tabcolsep}{5pt}
\renewcommand{\arraystretch}{1.2}
    \resizebox{.48\textwidth}{!}{%
    \begin{tabular}{ll||c|c|c|c||c}
    Training & Arch. & Kings & Old & Shop & St. Mary's & Street \\ 
    \hline
    \multirow{2}{*}{Real} & MLP & 0.46/0.72 & 2.19/2.75 & 0.55/2.06 & 1.48/3.35 & -  \\
     & Trans. & 0.45/0.63 & 1.62/2.60 & 0.54/2.05 & 1.28/2.99 &  -  \\ 
    \hdashline[4pt/1.61803398875pt]
    \rule{0pt}{.9\normalbaselineskip}
    \multirow{2}{1.5cm}{\raggedright Synthetic - \textit{In Dist.}} & MLP & 0.26/0.45 & \textbf{0.81}/1.37 & 0.29/1.16 & 0.73/1.95 & 6.0/18.6  \\
    & Trans. & \textbf{0.24}/\textbf{0.38} & 0.83/\textbf{1.05} & \textbf{0.26}/\textbf{1.06} & \textbf{0.58}/\textbf{1.56} &  \textbf{5.8}/\textbf{16.2} 
    \end{tabular}
    }
\end{center}
\vspace{-15pt}
\caption{Ablation study \wrt to different components of our localisation pipeline. We compare training with real image pairs only and our \oursA~ aproach. For each case, we also compare the use of MLP \vs Transformer.}%
\label{tab:ablation_main}
\vspace{-5pt}
\end{table}

\begin{table}[t!]
\begin{center}
\setlength{\tabcolsep}{2.5pt}
\renewcommand{\arraystretch}{1.4}
\scriptsize{
    \resizebox{.48\textwidth}{!}{%
    \begin{tabular}{cc||c||c|c|c||c}
        & & \multicolumn{5}{c}{\% of training images} \\
        Scene & \# images & All real & 10 & 25 & 50 & Out-of-Dist.  \\ 
        \hline
        Chess & 4000 & 0.13/5.25 &  0.11/4.32  & 0.08/2.88  & 0.06/2.15 & 0.03/1.10\\
        Heads & 1000 & 0.17/13.7 &  0.15/12.4 & 0.14/11.3 & 0.13/9.81 & 0.04/2.78\\
    \end{tabular}
    }%
}
\end{center}
\vspace{-15pt}
\caption{Results when using limited amount of data jointly with our synthetic images. We use \oursB~to generate a single novel view per training image for different amounts of data, and we use the resulting pairs for training. }%
\label{tab:ablation_proportion}
\vspace{-5pt}
\end{table}
For example, in \textit{Chess}, which contains 4000 images, the 50\% result means that 2000 real training images are selected, each paired with a synthetic view, adding up to a total of 4000 training images. For 25\% that would be 1000 real and synthetic images each for a total of 2000, and so on.
Table~\ref{tab:ablation_proportion} shows for both scenes that even when training with only 10\% of the original real images and their corresponding generated synthetic views, we obtain a higher performance than when using all the available real images. This result shows that the distribution of the relative poses is more important than the raw number of images used.

\parheader{MLP vs. Transformer.} 
\label{subsec:mlp_vs_transformer}
We also compare the MLP used in prior RPR methods~\cite{ding2019camnet,saha2018anchornet}, which predicts the relative pose from two concatenated image embeddings, to the Transformer we use in our \textit{Pose Regression Module}.
From Table~\ref{tab:ablation_main} we see that although the overall performance boost of our localisation pipeline can be largely  attributed to training with synthetic views, using the Transformer still gives a notable improvement of the median errors.

\parheader{Quality of synthetic images.} 
\label{subsec:quality_of_synthetic_images}
In Figure~\ref{fig:synth_images} we show examples of synthetic images we use for training.
Synthetic images from the MVS depths only,  suffer greatly from its sparsity, showing many holes and flying pixels.
However, the fusion of predicted dense and sparse MVS depths is able to produce much denser synthetic views. The lower accuracy of the predicted depth is manifested by flying pixels or stitching issues in some views, as shown in \eg the fourth and fifth columns in Figure~\ref{fig:synth_images}. The depth maps from the indoors scenes we use are denser, hence the effect of fusing the depths is less notable. The qualities of synthetic images vary, but even so, our network is capable of learning from such views and generalising to the real images in the test data. This is a significant observation, since it shows that due to the holistic nature of the pose regression methods, images with missing or noisy local regions, are still able to contribute to localisation. We include more examples of synthetic views in the supplementary material.
%
\section{Conclusion}
\label{sec:conclusion}
In this paper we have exposed and addressed the problem of mismatch between the training and test distributions for regressing camera pose. This is the most important factor responsible for the performance gap between the direct pose regression and structured methods. We proposed a strategy to generate new camera poses guided by the probabilistic distribution of the training poses to correct the data bias.  We have introduced a novel training process that is based on new view synthesis by combining dense scale-agnostic depth with sparse metric depth. 
We also adapt the Transformer model and show its superiority to the previously used MLP for pose regression. 

Importantly, the presented improvements bring the performance of the regression methods closer to or higher than the structure-based methods, which  is a significant step towards an end-to-end approach that can eventually replace the structured ones. Our approach is the first direct regression method to achieve better results than image retrieval in very challenging large outdoor scenes.

\parheader{Acknowledgement.} This research was supported by UK EPSRC project EP/S032398/1 and EP/N007743/1.

{\small
\bibliographystyle{ieee_fullname}
\bibliography{references}

\begin{thebibliography}{10}\itemsep=-1pt

\bibitem{arandjelovic2016netvlad}
Relja Arandjelovi{\'c}, Petr Gronat, Akihiko Torii, Tomas Pajdla, and Josef
  Sivic.
\newblock Net{VLAD}: {CNN} architecture for weakly supervised place
  recognition.
\newblock In {\em CVPR}, 2016.

\bibitem{arandjelovic2014dislocation}
Relja Arandjelovi{\'c} and Andrew Zisserman.
\newblock {DisLocation}: Scalable descriptor distinctiveness for location
  recognition.
\newblock In {\em ACCV}, 2014.

\bibitem{balntas2018relocnet}
Vassileios Balntas, Shuda Li, and Victor~Adrian Prisacariu.
\newblock {RelocNet}: Continuous metric learning relocalisation using neural
  nets.
\newblock In {\em ECCV}, 2018.

\bibitem{brachmann2018dsac}
Eric Brachmann, Alexander Krull, Sebastian Nowozin, Jamie Shotton, Frank
  Michel, Stefan Gumhold, and Carsten Rother.
\newblock {DSAC} - differentiable ransac for camera localization.
\newblock In {\em CVPR}, 2017.

\bibitem{brachmann2018dsac++}
Eric Brachmann and Carsten Rother.
\newblock Learning less is more - {6D} camera localization via {3D} surface
  regression.
\newblock In {\em CVPR}, 2018.

\bibitem{brachmann2020dsacstar}
Eric Brachmann and Carsten Rother.
\newblock Visual camera re-localization from {RGB} and {RGB-D} images using
  {DSAC}.
\newblock {\em arXiv}, 2020.

\bibitem{brahmbhatt2018mapnet}
Samarth Brahmbhatt, Jinwei Gu, Kihwan Kim, James Hays, and Jan Kautz.
\newblock Geometry-aware learning of maps for camera localization.
\newblock In {\em CVPR}, 2018.

\bibitem{chidlovskii2020adversarial}
Boris Chidlovskii and Assem Sadek.
\newblock Adversarial transfer of pose estimation regression.
\newblock In {\em ECCV Workshops}, 2020.

\bibitem{extremeview}
Inchang Choi, Orazio Gallo, Alejandro Troccoli, Min~H Kim, and Jan Kautz.
\newblock Extreme view synthesis.
\newblock In {\em ICCV}, 2019.

\bibitem{deng2009ImageNet}
Jia Deng, Wei Dong, Richard Socher, Li-Jia Li, {Kai Li}, and {Li Fei-Fei}.
\newblock Imagenet: A large-scale hierarchical image database.
\newblock In {\em CVPR}, 2009.

\bibitem{ding2019camnet}
Mingyu Ding, Zhe Wang, Jiankai Sun, Jianping Shi, and Ping Luo.
\newblock {CamNet}: Coarse-to-fine retrieval for camera re-localization.
\newblock In {\em ICCV}, 2019.

\bibitem{dosovitskiy2021an}
Alexey Dosovitskiy, Lucas Beyer, Alexander Kolesnikov, Dirk Weissenborn,
  Xiaohua Zhai, Thomas Unterthiner, Mostafa Dehghani, Matthias Minderer, Georg
  Heigold, Sylvain Gelly, Jakob Uszkoreit, and Neil Houlsby.
\newblock An image is worth 16x16 words: Transformers for image recognition at
  scale.
\newblock In {\em International Conference on Learning Representations}, 2021.

\bibitem{dusmanu2019d2net}
Mihai Dusmanu, Ignacio Rocco, Tomas Pajdla, Marc Pollefeys, Josef Sivic,
  Akihiko Torii, and Torsten Sattler.
\newblock {D2-Net}: A trainable cnn for joint detection and description of
  local features.
\newblock In {\em CVPR}, 2019.

\bibitem{fischler1981ransac}
Martin~A. Fischler and Robert~C. Bolles.
\newblock Random sample consensus: A paradigm for model fitting with
  applications to image analysis and automated cartography.
\newblock {\em Commun. ACM}, 24(6):381--395, June 1981.

\bibitem{gao2003p3p}
Xiao-Shan {Gao}, Xiao-Rong {Hou}, Jianlian {Tang}, and Hang-Fei {Cheng}.
\newblock Complete solution classification for the perspective-three-point
  problem.
\newblock {\em IEEE Transactions on Pattern Analysis and Machine Intelligence},
  25(8):930--943, Aug 2003.

\bibitem{glocker20137scenes}
Ben Glocker, Shahram Izadi, Jamie Shotton, and Antonio Criminisi.
\newblock Real-time {RGB-D} camera relocalization.
\newblock In {\em IEEE International Symposium on Mixed and Augmented Reality
  (ISMAR)}, October 2013.

\bibitem{he2016ResNet}
Kaiming He, Xiangyu Zhang, Shaoqing Ren, and Jian Sun.
\newblock Deep residual learning for image recognition.
\newblock In {\em CVPR}, 2016.

\bibitem{humenberger2020kapture}
Martin Humenberger, Yohann Cabon, Nicolas Guerin, Julien Morat, Jérôme
  Revaud, Philippe Rerole, Noé Pion, Cesar~Souza de, Vincent Leroy, and
  Gabriela Csurka.
\newblock Robust image retrieval-based visual localization using kapture, 2020.

\bibitem{kapture2020}
Martin Humenberger, Yohann Cabon, Nicolas Guerin, Julien Morat, Jérôme
  Revaud, Philippe Rerole, Noé Pion, Cesar de Souza, Vincent Leroy, and
  Gabriela Csurka.
\newblock Robust image retrieval-based visual localization using kapture, 2020.

\bibitem{jin2020image}
Yuhe Jin, Dmytro Mishkin, Anastasiia Mishchuk, Jiri Matas, Pascal Fua,
  Kwang~Moo Yi, and Eduard Trulls.
\newblock Image matching across wide baselines: From paper to practice, 2020.

\bibitem{kendall2016uncertainty}
Alex Kendall and Roberto Cipolla.
\newblock Modelling uncertainty in deep learning for camera relocalization.
\newblock In {\em ICRA}, 2016.

\bibitem{kendall2017posenet-geometric}
Alex Kendall and Roberto Cipolla.
\newblock Geometric loss functions for camera pose regression with deep
  learning.
\newblock In {\em CVPR}, 2017.

\bibitem{kendall2015posenet}
Alex Kendall, Matthew Grimes, and Roberto Cipolla.
\newblock {PoseNet}: A convolutional network for real-time {6-DOF} camera
  relocalization.
\newblock In {\em ICCV}, 2015.

\bibitem{kingma2015Adam}
Diederik~P. Kingma and Jimmy Ba.
\newblock Adam: A method for stochastic optimization.
\newblock In {\em ICLR}, 2015.

\bibitem{li2020hierarchical}
Xiaotian Li, Shuzhe Wang, Yi Zhao, Jakob Verbeek, and Juho Kannala.
\newblock Hierarchical scene coordinate classification and regression for
  visual localization.
\newblock In {\em Proceedings of the IEEE/CVF Conference on Computer Vision and
  Pattern Recognition}, pages 11983--11992, 2020.

\bibitem{li2018full}
Xiaotian Li, Juha Ylioinas, and Juho Kannala.
\newblock Full-frame scene coordinate regression for image-based localization.
\newblock In {\em Robotics: Science and Systems Conference}. University of
  Queensland, 2018.

\bibitem{liu2020neural}
Lingjie Liu, Jiatao Gu, Kyaw Zaw~Lin, Tat-Seng Chua, and Christian Theobalt.
\newblock Neural sparse voxel fields.
\newblock {\em Advances in Neural Information Processing Systems}, 33, 2020.

\bibitem{lowe2004sift}
David~G. Lowe.
\newblock Distinctive image features from scale-invariant keypoints.
\newblock In {\em IJCV}, 2004.

\bibitem{maddern2017robotcar}
Will Maddern, Geoff Pascoe, Chris Linegar, and Paul Newman.
\newblock {1 Year, 1000km: The Oxford RobotCar Dataset}.
\newblock {\em The International Journal of Robotics Research (IJRR)},
  36(1):3--15, 2017.

\bibitem{martinbrualla2021nerf-in-the-wild}
Ricardo Martin-Brualla, Noha Radwan, Mehdi S.~M. Sajjadi, Jonathan~T. Barron,
  Alexey Dosovitskiy, and Daniel Duckworth.
\newblock Nerf in the wild: Neural radiance fields for unconstrained photo
  collections.
\newblock In {\em CVPR}, 2021.

\bibitem{mildenhall2020nerf}
Ben Mildenhall, Pratul~P. Srinivasan, Matthew Tancik, Jonathan~T. Barron, Ravi
  Ramamoorthi, and Ren Ng.
\newblock Nerf: Representing scenes as neural radiance fields for view
  synthesis.
\newblock In {\em ECCV}, 2020.

\bibitem{mur-artal2016orb-slam}
Raul Mur-Artal, J.~M.~M. Montiel, and Juan~D. Tardos.
\newblock {ORB-SLAM}: A versatile and accurate monocular slam system.
\newblock {\em IEEE Transactions on Robotics}, 31(5):1147–1163, Oct 2015.

\bibitem{naseer2017deep}
Tayyab Naseer and Wolfram Burgard.
\newblock Deep regression for monocular camera-based {6-DoF} global
  localization in outdoor environments.
\newblock In {\em IROS}, 2017.

\bibitem{newcombe2011dtam}
Richard~A. Newcombe, Steven~J. Lovegrove, and Andrew~J. Davison.
\newblock {DTAM}: Dense tracking and mapping in real-time.
\newblock In {\em ICCV}, 2011.

\bibitem{ng2020solar}
Tony Ng, Vassileios Balntas, Yurun Tian, and Krystian Mikolajczyk.
\newblock {SOLAR}: Second-order loss and attention for image retrieval.
\newblock In {\em ECCV}, 2020.

\bibitem{niklaus20193dkensburn}
Simon Niklaus, Long Mai, Jimei Yang, and Feng Liu.
\newblock 3d ken burns effect from a single image.
\newblock In {\em TOG}, 2019.

\bibitem{noh2017delf}
Hyeonwoo Noh, Andr{\'e} Araujo, Jack Sim, Tobias Weyand, and Bohyung Han.
\newblock Image retrieval with deep local features and attention-based
  keypoints.
\newblock In {\em ICCV}, 2017.

\bibitem{pion2020benchmarking}
Noé Pion, Martin Humenberger, Gabriela Csurka, Yohann Cabon, and Torsten
  Sattler.
\newblock Benchmarking image retrieval for visual localization.
\newblock In {\em 3DV}, 2020.

\bibitem{radenovic2016gem}
Filip Radenovi{\'c}, Giorgos Tolias, and Ond{\u{r}}ej Chum.
\newblock {CNN} image retrieval learns from {BoW}: Unsupervised fine-tuning
  with hard examples.
\newblock In {\em ECCV}, 2016.

\bibitem{radenovic2018gem}
Filip Radenovi{\'c}, Giorgos Tolias, and Ond{\u{r}}ej Chum.
\newblock Fine-tuning {CNN} image retrieval with no human annotation.
\newblock {\em TPAMI}, 2018.

\bibitem{ranftl2020midas}
René Ranftl, Katrin Lasinger, David Hafner, Konrad Schindler, and Vladlen
  Koltun.
\newblock Towards robust monocular depth estimation: Mixing datasets for
  zero-shot cross-dataset transfer.
\newblock {\em TPAMI}, pages 1--1, 2020.

\bibitem{revaud2019aploss}
Jerome Revaud, Jon Almaz{\'{a}}n, Rafael Sampaio~de Rezende, and C{\'e}sar
  Roberto~de Souza.
\newblock Learning with average precision: Training image retrieval with a
  listwise loss.
\newblock In {\em ICCV}, 2019.

\bibitem{revaud2019r2d2}
Jerome Revaud, Philippe Weinzaepfel, César De~Souza, Noe Pion, Gabriela
  Csurka, Yohann Cabon, and Martin Humenberger.
\newblock {R2D2}: Repeatable and reliable detector and descriptor, 2019.

\bibitem{riba2020kornia}
Edgar Riba, Dmytro Mishkin, Daniel Ponsa, Ethan Rublee, and Gary Bradski.
\newblock Kornia: an open source differentiable computer vision library for
  pytorch.
\newblock In {\em Proceedings of the IEEE/CVF Winter Conference on Applications
  of Computer Vision}, pages 3674--3683, 2020.

\bibitem{riegler2020free}
Gernot Riegler and Vladlen Koltun.
\newblock Free view synthesis.
\newblock In {\em ECCV}, 2020.

\bibitem{saha2018anchornet}
Soham Saha, Girish Varma, and C.~V. Jawahar.
\newblock Improved visual relocalization by discovering anchor points.
\newblock In {\em BMVC}, 2018.

\bibitem{sarlin2018hfnet}
Paul{-}Edouard Sarlin, Cesar Cadena, Roland Siegwart, and Marcin Dymczyk.
\newblock From coarse to fine: Robust hierarchical localization at large scale.
\newblock In {\em CVPR}, 2019.

\bibitem{sarlin20superglue}
Paul-Edouard Sarlin, Daniel DeTone, Tomasz Malisiewicz, and Andrew Rabinovich.
\newblock {SuperGlue}: Learning feature matching with graph neural networks.
\newblock In {\em CVPR}, 2020.

\bibitem{sattler2016large-scale-burst}
Torsten Sattler, Michal Havlena, Konrad Schindler, and Marc Pollefeys.
\newblock Large-scale location recognition and the geometric burstiness
  problem.
\newblock In {\em CVPR}, 2016.

\bibitem{sattler2017active}
Torsten Sattler, Bastian Leibe, and Leif Kobbelt.
\newblock Efficient \& effective prioritized matching for large-scale
  image-based localization.
\newblock {\em TPAMI}, 39(9):1744--1756, 2017.

\bibitem{sattler2018benchmarking}
Torsten Sattler, Will Maddern, Carl Toft, Akihiko Torii, Lars Hammarstrand,
  Erik Stenborg, Daniel Safari, Masatoshi Okutomi, Marc Pollefeys, Josef Sivic,
  Fredrik Kahl, and Tomas Pajdla.
\newblock Benchmarking 6dof outdoor visual localization in changing conditions.
\newblock In {\em CVPR}, 2018.

\bibitem{sattler2017are-large-scale-3D}
Torsten {Sattler}, Akihiko {Torii}, Josef {Sivic}, Marc {Pollefeys}, Hajime
  {Taira}, Masatoshi {Okutomi}, and Tomas {Pajdla}.
\newblock Are large-scale 3d models really necessary for accurate visual
  localization?
\newblock In {\em CVPR}, 2017.

\bibitem{sattler2019understanding}
Torsten Sattler, Qunjie Zhou, Marc Pollefeys, and Laura-Taixe Leal.
\newblock Understanding the limitations of {CNN}-based absolute camera pose
  regression.
\newblock In {\em CVPR}, 2019.

\bibitem{schonberger2016colmap}
Johannes~L. Schönberger and Jan-Michael Frahm.
\newblock Structure-from-motion revisited.
\newblock In {\em CVPR}, 2016.

\bibitem{shotton2013scene}
Jamie Shotton, Ben Glocker, Christopher Zach, Shahram Izadi, Antonio Criminisi,
  and Andrew Fitzgibbon.
\newblock Scene coordinate regression forests for camera relocalization in
  {RGB-D} images.
\newblock In {\em CVPR}, 2013.

\bibitem{simonyan2015VGG}
Karen Simonyan and Andrew Zisserman.
\newblock Very deep convolutional networks for large-scale image recognition.
\newblock In {\em ICLR}, 2015.

\bibitem{srinivasan2019pushing}
Pratul~P. Srinivasan, Richard Tucker, Jonathan~T. Barron, Ravi Ramamoorthi, Ren
  Ng, and Noah Snavely.
\newblock Pushing the boundaries of view extrapolation with multiplane images.
\newblock In {\em CVPR}, 2019.

\bibitem{sweeney2015theia}
Chris Sweeney, Tobias Hollerer, and Matthew Turk.
\newblock Theia: A fast and scalable structure-from-motion library.
\newblock In {\em Proceedings of the 23rd ACM International Conference on
  Multimedia}, MM '15, page 693–696, 2015.

\bibitem{taira2018inloc}
Hajime Taira, Masatoshi Okutomi, Torsten Sattler, Mircea Cimpoi, Marc
  Pollefeys, Josef Sivic, Tomas Pajdla, and Akihiko Torii.
\newblock Inloc: Indoor visual localization with dense matching and view
  synthesis.
\newblock In {\em CVPR}, 2018.

\bibitem{tian2020hynet}
Yurun Tian, Axel Barroso-Laguna, Tony Ng, Vassileios Balntas, and Krystian
  Mikolajczyk.
\newblock {HyNet}: Learning local descriptor with hybrid similarity measure and
  triplet loss.
\newblock In {\em NeurIPS}, 2020.

\bibitem{torii201524-7}
Akihiko {Torii}, Relja {Arandjelović}, Josef {Sivic}, Masatoshi {Okutomi}, and
  Tomas {Pajdla}.
\newblock 24/7 place recognition by view synthesis.
\newblock In {\em CVPR}, 2015.

\bibitem{vaswani2017transformer}
Ashish Vaswani, Noam Shazeer, Niki Parmar, Jakob Uszkoreit, Llion Jones,
  Aidan~N. Gomez, Lukasz Kaiser, and Illia Polosukhin.
\newblock Attention is all you need.
\newblock In {\em NeurIPS}, 2017.

\bibitem{walch2017posenet-lstm}
Florian Walch, Caner Hazirbas, Laura Leal-Taixé, Torsten Sattler, Sebastian
  Hilsenbeck, and Daniel Cremers.
\newblock Image-based localization using {LSTM}s for structured feature
  correlation.
\newblock In {\em ICCV}, 2017.

\bibitem{wan2020boosting}
Yiming Wan, Wei Gao, Sheng Han, and Yihong Wu.
\newblock Boosting image-based localization via randomly geometric data
  augmentation.
\newblock In {\em 2020 IEEE International Conference on Image Processing
  (ICIP)}, pages 688--692. IEEE, 2020.

\bibitem{wang2020atloc}
Bing Wang, Changhao Chen, Chris~Xiaoxuan Lu, Peijun Zhao, Niki Trigoni, and
  Andrew Markham.
\newblock Atloc: Attention guided camera localization.
\newblock In {\em Proceedings of the AAAI Conference on Artificial
  Intelligence}, volume~34, pages 10393--10401, 2020.

\bibitem{xue2020multi-view}
Fei Xue, Xin Wu, Shaojun Cai, and Junqin Wang.
\newblock Learning multi-view camera relocalization with graph neural networks.
\newblock In {\em CVPR}, 2020.

\bibitem{zhang2020aachenv_1_1}
Zichao Zhang, Torsten Sattler, and Davide Scaramuzza.
\newblock Reference pose generation for long-term visual localization via
  learned features and view synthesis.
\newblock {\em International Journal of Computer Vision}, Dec 2020.

\bibitem{zhou2020kfnet}
Lei Zhou, Zixin Luo, Tianwei Shen, Jiahui Zhang, Mingmin Zhen, Yao Yao, Tian
  Fang, and Long Quan.
\newblock {KFNet}: Learning temporal camera relocalization using kalman
  filtering.
\newblock In {\em CVPR}, 2020.

\end{thebibliography}
}

\begin{figure*}[!ht]
\begin{center}
  \textbf{\Large Reassessing the Limitations of CNN Methods for Camera Pose Regression
  \vspace{5pt}
  \\ \large{Supplementary Material}
  }\\[0.8cm]
  
{ \large {Tony Ng$^{1*}$ \hspace{6pt}Adrian Lopez-Rodriguez$^{1*}$ \hspace{6pt}Vassileios Balntas$^{2}$   \hspace{6pt}
Krystian Mikolajczyk$^{1}$\\
$^1$Imperial College London \\
$^2$Facebook Reality Labs \\}
{\tt\small \{tony.ng14, adrian.lopez-rodriguez15, k.mikolajczyk\}@imperial.ac.uk}\\
{\tt\small vassileios@fb.com} \\
}
\end{center}
\end{figure*}
\FloatBarrier
\setcounter{section}{0}
\renewcommand{\thesection}{\arabic{section}}

\vspace{100pt}
\section{Extended implementation details}
\parheader{Pairs selection - real data.} We use directly the GeM~\cite{radenovic2016gem} rankings to form our pairs. Given a training image, we resize it to 180 pixels on the shorter side while maintaining the aspect ratio, and extract the GeM descriptor. The top $N$ Nearest Neighbours (NN) are retrieved by computing the dot product of the query to the remaining training images' GeM descriptors. The query and a randomly sampled NN from the top $N$ neighbours form our training pair. For outdoor data, we set $N=20$, meaning a training pair is randomly formed from a query and any of its 20 NNs. In indoor data, due to the smaller scale of the scenes we train on, we allow all of the images to form training pairs with each other. In \textit{Street}, where GeM descriptors have low precision, we remove those training pairs which camera position that have a L1-distance above 20 meters from the query. 
\parheader{In-Distribution.} The selection procedure for the training pairs follows the method described in \textit{Pairs Selection - Real Data}. After obtaining a pair of query-NN, we perturb the NN's camera pose with $\alpha_q=0.02$, $\alpha_t=\SI{1}{m}$ for outdoor data, whereas we set  $\alpha_q=0.02$, $\alpha_t=\SI{0.1}{m}$ for indoor data. The query is also augmented with a rotation-only perturbation of $\alpha_q=0.02$, \cf Section 3.2 in the paper for definitions of $\alpha_q$ and $\alpha_t$.
For both the NN and the query, we use a probability of 50\% of perturbing the poses and 50\% of using the original real poses and images.

\parheader{Out-of-Distribution.} Given a query image, we produce a synthetic pose by adding Gaussian perturbations to the query pose $C_q$. For the rotation perturbation, we use the standard yaw, roll and pitch axes definitions. For each of these axes, we choose an angle $\phi$ by sampling from a Gaussian distribution, \eg, for the yaw axis we select an angle $\phi_y$ by sampling from $\mathcal{N}\left(\textbf{0},\sigma_y\right)$. We then transform these sampled angles $\phi_y, \phi_r, \phi_p$ to a rotation matrix and use it to transform the original rotation in $C_q$ to a new synthetic rotation. For the translations, we follow a two-stage approach. We first perturb the original translation in $C_q$ with a larger perturbation sampled from a Gaussian with standard deviation $\alpha_{t,large}$ to form our perturbed translation. To avoid situation in which after this large scale perturbation our pose is behind a wall, or any other \textit{out-of-bounds} situation, we select the closest training pose to the perturbed translation and, to this closest training pose, we apply a smaller translation perturbation with standard deviation $\alpha_{t,small}$ to form our final synthetic translation. We select the standard deviations of the perturbations we use by inspecting the training data. For outdoor data, we notice that most of the rotations are around the yaw axis, hence we set our the standard deviations of our rotation perturbations to $\sigma_y=30^\circ$ around the yaw axis, $\sigma_r=10^\circ$ around the roll axis and $\sigma_p=2.5^\circ$ around the pitch axis. For the translation, we set $\alpha_{t,large}=\SI{10}{m}$ in the ground-plane, $\SI{0.1}{m}$ in the height axis, and $\alpha_{t,small}$ to $\SI{0.5}{m}$ in the ground plane and $\SI{0.1}{m}$ in the height axis. For indoor data, we assume the distribution of rotations is independent of the axis, hence we use a standard deviation of $15^\circ$ in all directions, \ie, $\sigma_y=\sigma_r=\sigma_p=15^\circ$. For the indoor translation perturbations, we set $\alpha_{t,large}=\SI{0.5}{m}$ and $\alpha_{t,large}=\SI{0.25}{m}$ in all directions. After we have created our synthetic NN pose by perturbing the query pose, we synthesise the new view by selecting the closest 10 images in translation (among those that are within a rotation threshold of $15^\circ/30^\circ$ for indoor/outdoor data) in the training set as the source images. We generate synthetic views with these, iteratively project from the depth maps of the NNs until either 10 images are sourced or 80\% of the synthetic image's pixels are filled. For each element in the batch, we randomly select with a probability of 25\% the real NN chosen following the method explained in \textit{Pairs Selection - Real Data}, instead of generating synthetically the NN. We also augment the real query image using the same rotation perturbation used in the \textit{In-Distribution} approach, with $\alpha_q=0.02$. As some of the generated images contain few projected pixels, we do not compute any loss for those images with less than 30\% of pixels projected. During the first 700 epochs of training, we randomly assign the synthetic view as either the NN or query with a probability of 50\%, whereas we fix the synthetic view as the NN after 700 epochs, which we found experimentally to increase the performance.

\begin{table}[t!]
\begin{center}
\setlength{\tabcolsep}{2.5pt}
\small{
    \begin{tabular}{c|cc}
        Method & St. Mary's & Street  \\ 
        \hline
        GeM~\cite{radenovic2016gem} & 3.01/9.37 & 6.39/23.6 \\
        GeM+Out-of-Distribution & 0.47/1.23 & 4.33/9.72\\
        \hline
        DenseVLAD~\cite{torii201524-7} & 2.60/7.83  & 5.85/23.2 \\
        DenseVLAD+Out-of-Distribution & 0.46/1.20 & 4.19/9.15\\
    \end{tabular}
}
\end{center}
\vspace{-5pt}
\caption{Comparison of results in two scenes of the \textit{Cambridge Landmarks}~\cite{kendall2015posenet} dataset when using DenseVLAD and GeM to retrieve our NN at inference time. Reported results are the median translation (m) / rotation ($^\circ$) errors. }%
\label{tab:densevlad}
\vspace{0pt}
\end{table}

\section{Results with DenseVLAD}
Table \ref{tab:densevlad} shows the results of our method when using the DenseVLAD~\cite{torii201524-7} top-1 NN at inference time compared to when using the top-1 GeM NN. We used the official MATLAB\footnote{\url{http://www.ok.ctrl.titech.ac.jp/~torii/project/247/}} code for the DenseVLAD implementation. The slight difference between our results and the DenseVLAD results reported in~\cite{sattler2019understanding} is due to the default settings in the official DenseVLAD code, which we adopt, aggregates from multiscale SIFT~\cite{lowe2004sift}. However, in~\cite{sattler2019understanding}, single-scale SIFT was used to compute DenseVLAD which was found to be more suitable for localisation. We use the same pose regression model and weights for both the GeM and the DenseVLAD results. The results in Table~\ref{tab:densevlad} show that having better NNs at inference alone improves the relative pose predictions of our model without needing any further training.
\section{Detailed architecture}
\begin{figure}[!t]
    \begin{center}
        \includegraphics[width=1.\linewidth, trim={
        5pt, 5pt, 0pt, 0pt}, clip]{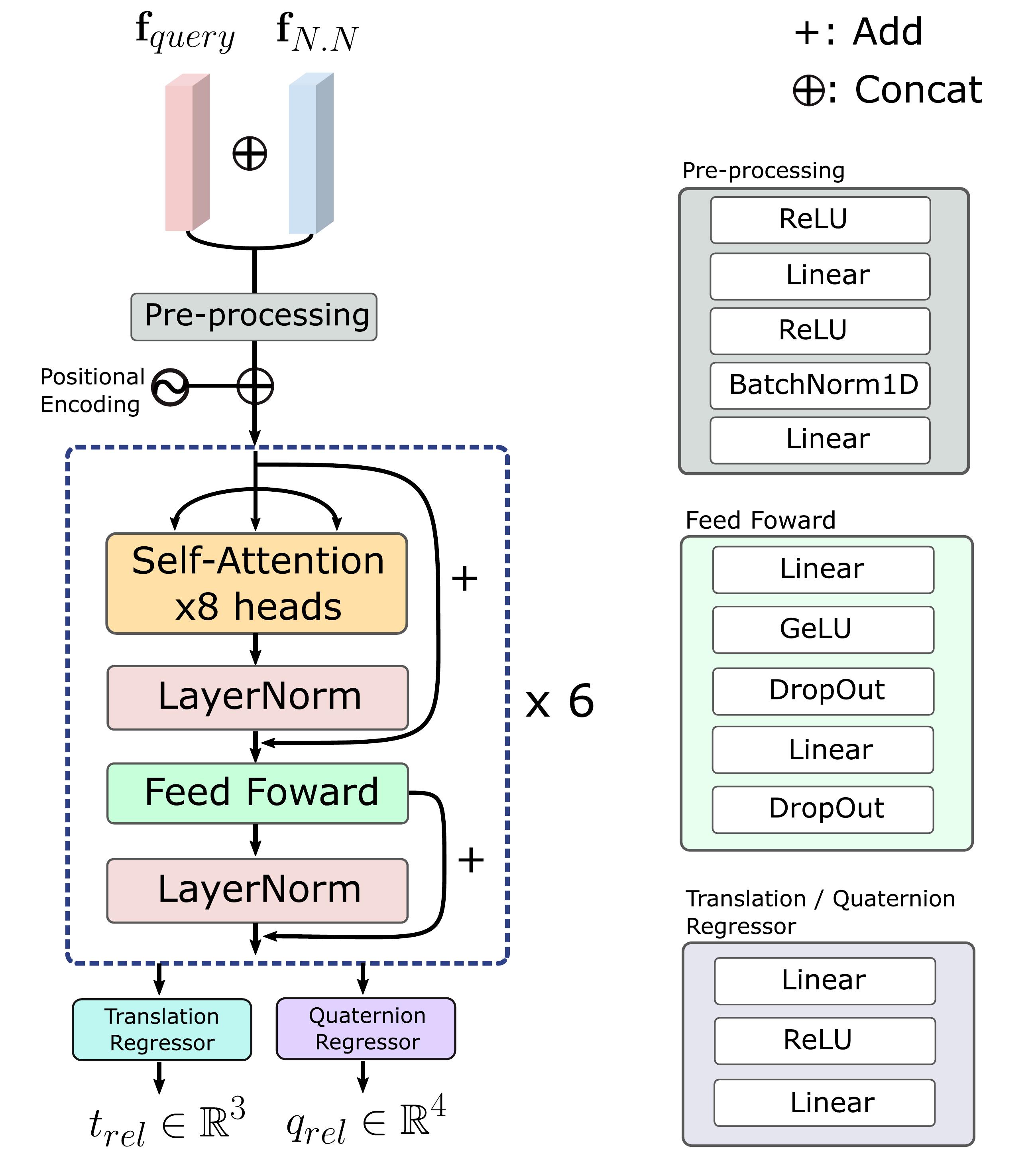}
    \end{center}
    \vspace{-5pt}
    \caption{Architecture of our Pose Regression Transformer.}
    \label{fig:transformer}
    \vspace{-5pt}
\end{figure}
Our base CNN for the relative regression model is a ResNet-34~\cite{he2016ResNet} pretrained in ImageNet~\cite{deng2009ImageNet}. The feature embedding of dimensionality 512 output by the ResNet-34 model is passed to our \textit{Relative Pose Regression Transformer}~\cite{vaswani2017transformer}, which is shown in Figure \ref{fig:transformer}. Our transformer encoder receives a sequence of two embeddings of dimensionality 512, the query $f_{query}$ and the NN embedding $f_{N.N.}$, which are first passed to a \textit{Pre-processing} module (\texttt{[Linear(512,2048), ReLU, BatchNorm1D, Linear(2048, 512)]}) before being fed to a transformer encoder that contains 6 layers and 8 heads, where each head has a dimensionality of 128. The dropout used in the \textit{Feed Forward} modules is set to 0.1. Before the two vectors are fed to the transformer encoder, we concatenate to both of them a positional encoding of size 128 that allows the transformer to know which embedding comes from the query and which from the NN. Thus, our transformer encoder receives embeddings of size $512+128=640$. The transformer output corresponding to the position of the NN is then fed to two different regressors, the \textit{Translation Regressor} (\texttt{[Linear(640,640), ReLU, Linear(640, 3)]}) and the \textit{Quaternion Regressor} (\texttt{[Linear(640,640), ReLU, Linear(640, 4)]}). Due to issues with batch normalisation, we switch the networks to evaluation mode after the first learning rate drop at 500 epochs.
\begin{table}[t!]
\scriptsize{
\setlength{\tabcolsep}{3pt}
\renewcommand{\arraystretch}{1.1}
\begin{center}
\resizebox{.475\textwidth}{!}{%
\begin{tabular}{cl|c|c|c|c|c|c|c}
& & \multicolumn{7}{c}{\footnotesize{7 Scenes}} \\
 & & Chess & Fire & Heads & Office & Pumpkin & Kitchen & Stairs  \\ \hline

\hline

\rule{0pt}{.9\normalbaselineskip}

& Real & 0.13/5.25 & 0.32/13.4  & 0.17/13.7 & 0.24/7.10   & 0.26/6.25 &  0.27/6.77 & 0.30/13.3  \\
& O.o.D.+Random & 0.03/1.20 & 0.04/2.07 & 0.05/3.37 & 0.04/1.41 & 0.04/1.24 & 0.06/1.71 & 0.09/1.79 \\
& O.o.D.+GeM & 0.03/1.10 & 0.04/1.64 &  0.04/2.78 & 0.04/1.35 & 0.04/1.20 & 0.05/1.62 & 0.10/1.62 
 
\end{tabular}
}
\end{center}
}%
\caption{Additional quantitative results in 7 Scenes~\cite{glocker20137scenes}. \textit{Real} refers to training with only real data, and we include results for our \oursB~method (\textit{O.o.D}) when using during inference either the GeM top-1 NN or a randomly chosen database image as our NN. Reported results are the median translation (m) / rotation ($^\circ$) errors.}%
\label{tab:results_random}
\vspace{0pt}
\end{table}
\section{Additional results in 7 Scenes}
We include additional results for \textit{7 Scenes}~\cite{glocker20137scenes} in Table \ref{tab:results_random}. \textit{Real} shows the performance when training only with real pairs. Due to the scenes available in \textit{7 Scenes} being constrained to a small region, we used a training strategy in which we allowed all of the available real images to form training pairs with each other. For real images only, our network overfits quickly to the training set, generalising poorly to the test set as the poses of the training images come from a limited number of video sequences. Our \textit{Out-of-Distribution} method, which forms training pairs by using a Gaussian perturbation based approach that allows the network to learn from a less biased relative, improves upon the \textit{Real} results by a large margin in all cases as we see in \textit{O.o.D.+GeM} in Table \ref{tab:results_random}. As with the \oursB~approach we learn to predict poses from a wide range of distances and rotations, we explore in the line \textit{O.o.D.+Random} in Table \ref{tab:results_random} the impact of using a randomly chosen Nearest Neighbour during inference instead of using the top-1 GeM NN. Table \ref{tab:results_random} 
shows that the performance is similar regardless of the NN image used, which shows that our method is robust to the choice of NN in smaller scenes.

\section{Inference speed analysis}
We use for our experiments a NVIDIA Titan RTX GPU, 64GB of RAM, a NVMe SSD, and a i9-10980XE CPU at 3.00 GHz. We assume that the embeddings used for retrieval for the training images are already precomputed, and during inference we only need to compute the GeM embedding for the query images. Similarly, we also assume our $f_{N.N}$, which are shown in Figure~\ref{fig:transformer}, are also precomputed. For these tests we use a batch size of 16, and unless stated otherwise we ran the tests in the \textit{Cambridge Landmarks}~\cite{kendall2015posenet} dataset with input image size of $180\times320$.

\parheader{Retrieval.} We use a pretrained GeM ResNet-50 network for retrieval. The inference time for the retrieval network for a batch of 16 images of size $180\times320$ is $\SI{21}{ms}$, or 762 images/second.

\parheader{Relative pose regression.} Our relative pose estimation network is capable of doing inference in a batch of 16 images of size $180\times320$ in only $\SI{20}{ms}$, or 800 images/second. We assume the features for the database images are already precomputed.

\parheader{Total.} Our final method needs two forward passes, one for the retrieval network, and one for the relative pose estimation network. Hence, our total computational time per batch is of $\SI{41}{ms}$, which translates to 390 images per second. However, to the given estimates we need to add the effects of loading the data, searching for the NN and the transformation from relative to absolute pose. The whole pipeline computes the poses for the 343 images in the test set of \textit{KingsCollege} (\textit{Cambridge Landmarks} dataset) in $\SI{1.10}{s}$ using a dataloader with 16 workers, amounting to 310 images/second. We benefit from parallelisation, as when using a batch size of 1 we achieve a speed of $\sim$50 imgs/s. For \textit{7 Scenes}~\cite{glocker20137scenes}, where we use an image resolution of $180\times 240$, we compute the 1000 test poses in \textit{7Scenes-heads} in $\SI{2.75}{s}$, amounting to 364 images/second. Additionally, in \textit{7 Scenes}, we obtain similar results by choosing a random database image as NN (\ie, skipping the retrieval step) as shown in Table \ref{tab:results_random}, which makes our method capable of computing the poses in \textit{heads} in $\SI{1.90}{s}$, or 526 images/second.

\parheader{Comparison.} Our inference time shows that our method is fast compared to 3d-based methods. Even though the hardware is different, in DSAC++~\cite{brachmann2018dsac++} computing a query pose takes $\SI{200}{ms}$ per image, or 5 images/second. ActiveSearch~\cite{sattler2017active} is capable of computing the pose for a query image in $\SI{36}{ms}$, or 28 images/second. In both cases our method is considerably faster. Compared to the \textit{7 Scenes} state-of-the-art, CamNet~\cite{ding2019camnet}, their base approach uses two relative pose predictions (coarse+fine) from two different NNs, and their final results include a RANSAC approach applied to the pose predictions that requires predicting the relative pose from 5 different NNs.

\section{Qualitative examples of synthetic views}
\begin{figure*}[t]
    \centering
    \begin{subfigure}[t]{\textwidth}
    \includegraphics[trim={0cm 0cm 0 0},width=\linewidth,clip]{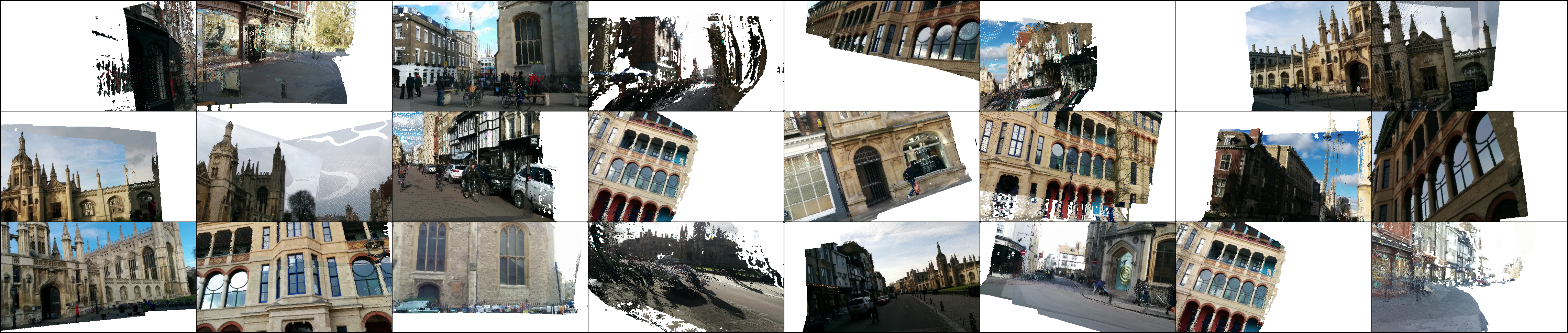}
    \caption{Synthetic images generated by using our \oursB~approach in outdoor scenes}
    \end{subfigure}
    \begin{subfigure}[t]{\textwidth}
    \includegraphics[trim={0cm 0cm 0 0},width=\linewidth,clip]{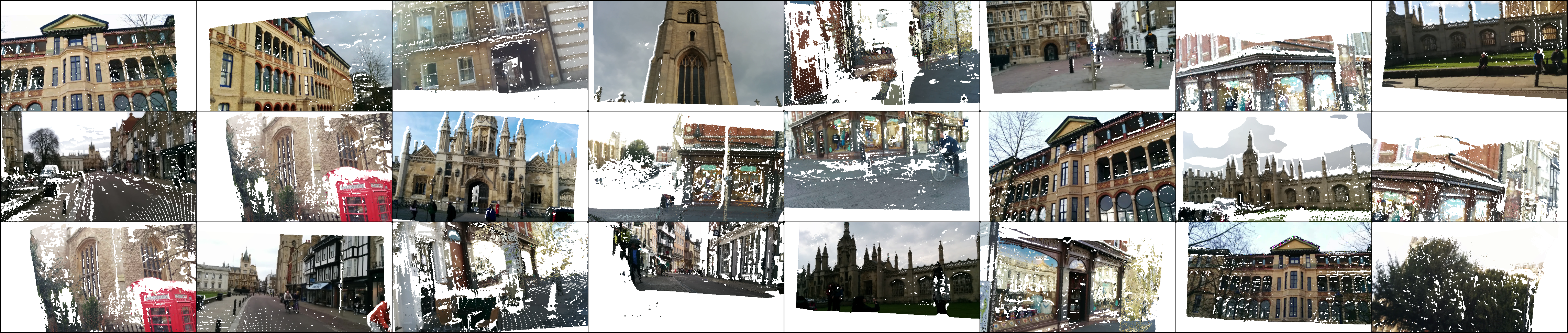}
    \caption{Synthetic images by using our \oursA~approach in outdoor scenes}
    \end{subfigure}
    \begin{subfigure}[t]{\textwidth}
    \includegraphics[trim={0cm 0cm 0 0},width=\linewidth,clip]{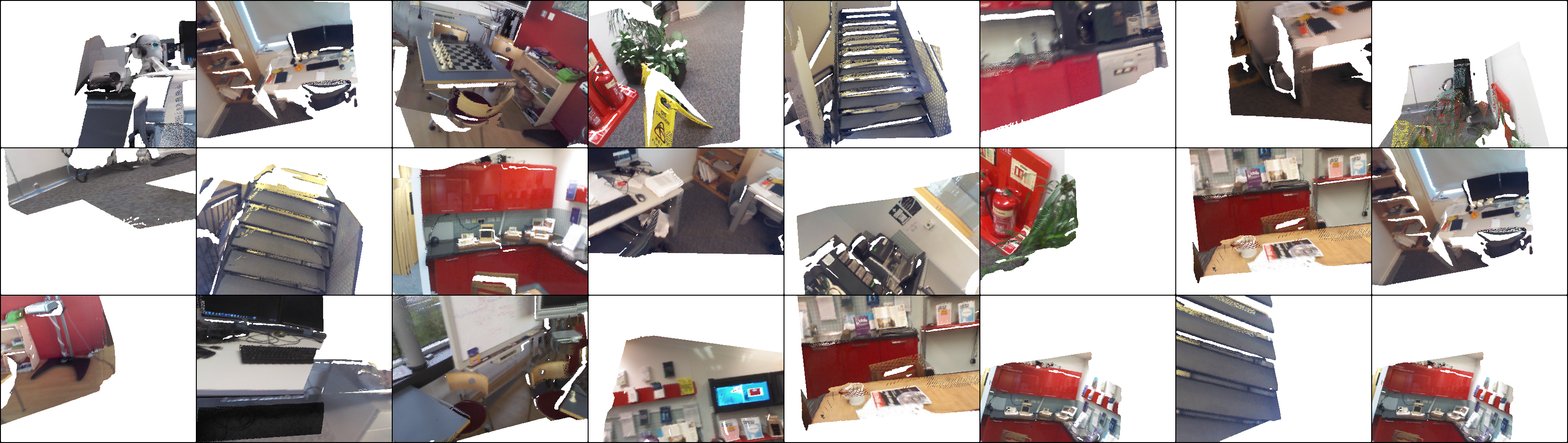}
    \caption{Synthetic images generated by using our \oursB~approach in indoor scenes}
    \end{subfigure}
    \begin{subfigure}[t]{\textwidth}
    \includegraphics[trim={0cm 0cm 0 0},width=\linewidth,clip]{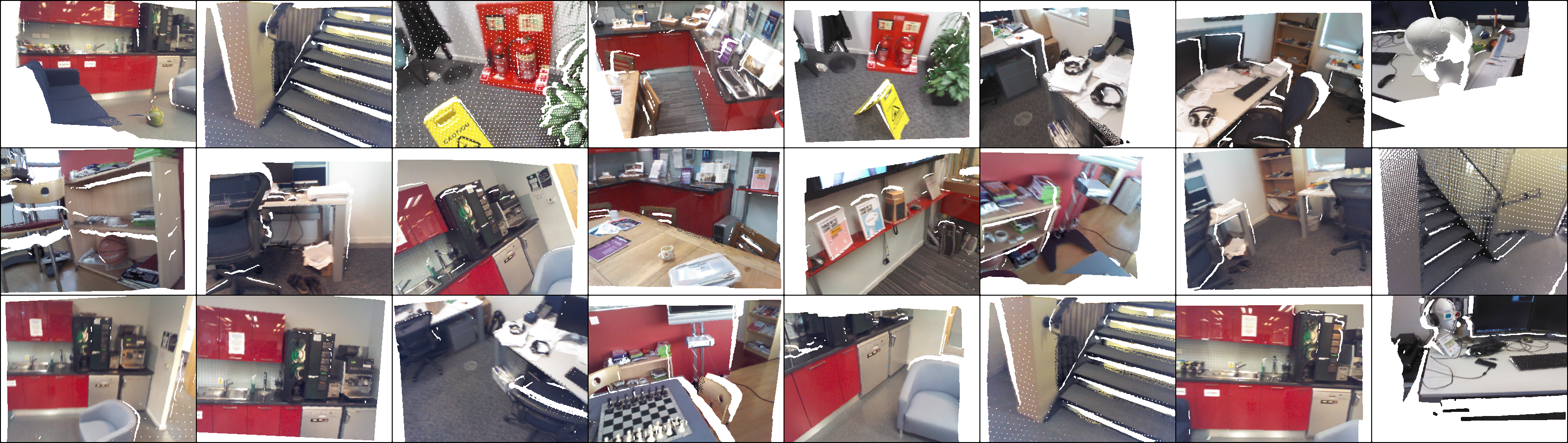}
    \caption{Synthetic images generated by using our \oursA~approach in indoor scenes}
    \end{subfigure}
    \caption{Synthetic images we use to train our relative pose regressor. The images are randomly selected directly from our method, we only filtered those images containing less than 30\% non-blank pixels.}
    \label{fig:synth_images2}
\end{figure*}

Figure \ref{fig:synth_images2} shows some examples of synthetic views for both indoor and outdoor data for our \oursB~approach. We filtered the images with less than 30\% of projected pixels as they are not taken into account to compute the loss. The indoor synthetic views, due to the larger density of views and the more constrained scenes, present usually a higher quality of synthesised views.

\end{document}